\documentclass[nohyperref]{article}

\usepackage{microtype}
\usepackage{graphicx}
\usepackage{subfig}
\usepackage{booktabs} %

\usepackage[colorlinks,citecolor=darkgreen,linkcolor=firebrick,urlcolor=firebrick]{hyperref}

\usepackage[utf8]{inputenc}
\usepackage{amsmath}
\usepackage{amssymb}
\usepackage{mathtools}
\numberwithin{equation}{section}
\usepackage{slashed}
\usepackage{braket}
\usepackage{enumitem}
\usepackage[bottom]{footmisc}
\usepackage{graphicx}
\usepackage{bm}
\usepackage{subfig}
\usepackage{physics}
\usepackage{xcolor}
\usepackage{natbib}
\usepackage{mdframed}
\usepackage{diagbox}
\usepackage{appendix}
\usepackage{titletoc}
\usepackage{etoc}
\usepackage{caption}

\usepackage[capitalize,noabbrev]{cleveref}

\usepackage{lipsum}
\usepackage{titlesec}
\usepackage{wrapfig,lipsum,booktabs}
\titlespacing\section{0pt}{2pt plus 2pt minus 1pt}{-1pt plus 1pt minus 1pt}
\titlespacing\subsection{0pt}{1pt plus 2pt minus 1pt}{-1pt plus 1pt minus 1pt}
\titlespacing\subsubsection{0pt}{1pt plus 2pt minus 1pt}{-1pt plus 1pt minus 1pt}

\usepackage{array}
\usepackage{makecell}
\usepackage{multirow}

\usepackage{CJK}

\renewenvironment{figure}[1][]{
  \begin{originalfigure}[#1]
    \begin{mdframed}[linecolor=black!0,backgroundcolor=black!1]
}{
    \end{mdframed}
  \end{originalfigure}
}

\def\half{{\frac{1}{2}}}

\def\0{{(0)}}
\def\1{{(1)}}
\def\2{{(2)}}

\def\<{\langle }
\def\>{\rangle }

\newcommand{\bea}{\begin{eqnarray}}
\newcommand{\eea}{\end{eqnarray}}

\usepackage{slashed}

\def\({\left(}
\def\){\right)}
\def\[{\left[}
\def\]{\right]}

\usepackage{dashbox}
\usepackage{xcolor}
\usepackage{colortbl}
\usepackage{arydshln}
\definecolor{lightyellow}{rgb}{1.0, 0.95, 0.7}
\definecolor{Blue}{rgb}{0.0, 0.4, 1.0}
\definecolor{blue}{rgb}{0,0,1}
\definecolor{darkgreen}{rgb}{0.,0.6,0.}

\definecolor{colorA}{rgb}{1,0,0}
\definecolor{colorB}{rgb}{0,0.3,1}
\definecolor{colorC}{rgb}{0.9,0.8,0.2}
\definecolor{colorD}{rgb}{0,0.65,0}
\definecolor{lesslightgray}{rgb}{0.5,0.5,0.5}
\definecolor{darkgreen}{rgb}{0,0.40,0}
\definecolor{firebrick}{rgb}{0.698,0.133,0.133}

\definecolor{light-gray}{gray}{0.95}
\newcommand{\code}[1]{\colorbox{light-gray}{\footnotesize\texttt{#1}}}

\let\hat\widehat

\newcommand{\calC}{\mathcal{C}}
\newcommand{\calD}{\mathcal{D}}
\newcommand{\calE}{\mathcal{E}}

\newcommand{\calG}{\mathcal{G}}

\newcommand{\calL}{\mathcal{L}}

\newcommand{\calN}{\mathcal{N}}

\newcommand{\calV}{\mathcal{V}}

\newcommand{\calZ}{\mathcal{Z}}

\newcommand{\bX}{\textbf{X}}
\newcommand{\bY}{\textbf{Y}}

\newcommand{\ba}{\bm{a}}

\newcommand{\bp}{\bm{p}}

\newcommand{\bx}{\textbf{x}}
\newcommand{\by}{\textbf{y}}

\newcommand{\sT}{ \mathsf{T} }

\newcommand{\E}{\mathop{\mathbb{E}}}

\newcommand{\sumN}{\sum_{i=1}^N}

\usepackage{amsthm}
\usepackage{nicefrac}

\usepackage{lipsum}

\makeatletter
\let\save@mathaccent\mathaccent
\newcommand*\if@single[3]{%
    \setbox0\hbox{${\mathaccent"0362{#1}}^H$}%
    \setbox2\hbox{${\mathaccent"0362{\kern0pt#1}}^H$}%
    \ifdim\ht0=\ht2 #3\else #2\fi
}
\newcommand*\rel@kern[1]{\kern#1\dimexpr\macc@kerna}
\newcommand*\widebar[1]{\@ifnextchar^{{\wide@bar{#1}{0}}}{\wide@bar{#1}{1}}}
\newcommand*\wide@bar[2]{\if@single{#1}{\wide@bar@{#1}{#2}{1}}{\wide@bar@{#1}{#2}{2}}}
\newcommand*\wide@bar@[3]{%
    \begingroup
    \def\mathaccent##1##2{%
        \let\mathaccent\save@mathaccent
        \if#32 \let\macc@nucleus\first@char \fi
        \setbox\z@\hbox{$\macc@style{\macc@nucleus}_{}$}%
        \setbox\tw@\hbox{$\macc@style{\macc@nucleus}{}_{}$}%
        \dimen@\wd\tw@
        \advance\dimen@-\wd\z@
        \divide\dimen@ 3
        \@tempdima\wd\tw@
        \advance\@tempdima-\scriptspace
        \divide\@tempdima 10
        \advance\dimen@-\@tempdima
        \ifdim\dimen@>\z@ \dimen@0pt\fi
        \rel@kern{0.6}\kern-\dimen@
        \if#31
        \overline{\rel@kern{-0.6}\kern\dimen@\macc@nucleus\rel@kern{0.4}\kern\dimen@}%
        \advance\dimen@0.4\dimexpr\macc@kerna
        \let\final@kern#2%
        \ifdim\dimen@<\z@ \let\final@kern1\fi
        \if\final@kern1 \kern-\dimen@\fi
        \else
        \overline{\rel@kern{-0.6}\kern\dimen@#1}%
        \fi
    }%
    \macc@depth\@ne
    \let\math@bgroup\@empty \let\math@egroup\macc@set@skewchar
    \mathsurround\z@ \frozen@everymath{\mathgroup\macc@group\relax}%
    \macc@set@skewchar\relax
    \let\mathaccentV\macc@nested@a
    \if#31
    \macc@nested@a\relax111{#1}%
    \else
    \def\gobble@till@marker##1\endmarker{}%
    \futurelet\first@char\gobble@till@marker#1\endmarker
    \ifcat\noexpand\first@char A\else
    \def\first@char{}%
    \fi
    \macc@nested@a\relax111{\first@char}%
    \fi
    \endgroup
    }
\makeatother

\usepackage{titletoc}

\usepackage[accepted]{icml2023}

\usepackage{amsmath}
\usepackage{amssymb}
\usepackage{mathtools}
\usepackage{amsthm}

\newtheorem{corollary}{Corollary}[section]

\usepackage{amsthm}
\makeatletter
\def\th@remark{%
  \thm@headfont{\bfseries}%
  \normalfont %
  \thm@preskip\topsep \divide\thm@preskip\tw@
  \thm@postskip\thm@preskip
}
\makeatother
\usepackage[many]{tcolorbox}
\theoremstyle{definition}
\newtheorem{theorem}{Theorem}[section]
\tcolorboxenvironment{theorem}{
  breakable,
  breakable,
  colback=black!10,
  colframe=white,%
  width=\dimexpr\linewidth+10pt\relax,%
  enlarge left by=-5pt,%
  enlarge right by=-5pt,%
  boxsep=5pt,%
  boxrule=0pt,
  left=0pt,right=0pt,top=0pt,bottom=0pt,
  sharp corners,
  before skip=\topsep,
  after skip=\topsep
}
\newtheorem{lemma}{Lemma}[section]
\tcolorboxenvironment{lemma}{
  breakable,
  colback=black!10,
  colframe=white,%
  width=\dimexpr\linewidth+10pt\relax,%
  enlarge left by=-5pt,%
  enlarge right by=-5pt,%
  boxsep=5pt,%
  boxrule=0pt,
  left=0pt,right=0pt,top=0pt,bottom=0pt,
  sharp corners,
  before skip=\topsep,
  after skip=\topsep
}

\theoremstyle{definition}
\newtheorem{definition}{Definition}[section]
\tcolorboxenvironment{definition}{
  breakable,
  colback=black!10,
  colframe=white,%
  width=\dimexpr\linewidth+10pt\relax,%
  enlarge left by=-5pt,%
  enlarge right by=-5pt,%
  boxsep=5pt,%
  boxrule=0pt,
  left=0pt,right=0pt,top=0pt,bottom=0pt,
  sharp corners,
  before skip=\topsep,
  after skip=\topsep
}
\theoremstyle{remark}
\newtheorem{remark}{Remark}[section]
\newtheorem{assumption}{Assumption}[section]
\tcolorboxenvironment{assumption}{
  breakable,
  colback=black!10,
  colframe=white,%
  width=\dimexpr\linewidth+10pt\relax,%
  enlarge left by=-5pt,%
  enlarge right by=-5pt,%
  boxsep=5pt,%
  boxrule=0pt,
  left=0pt,right=0pt,top=0pt,bottom=0pt,
  sharp corners,
  before skip=\topsep,
  after skip=\topsep
}

\tcolorboxenvironment{problem}{
  breakable,
  colback=black!10,
  colframe=white,%
  width=\dimexpr\linewidth+10pt\relax,%
  enlarge left by=-5pt,%
  enlarge right by=-5pt,%
  boxsep=5pt,%
  boxrule=0pt,
  left=0pt,right=0pt,top=0pt,bottom=0pt,
  sharp corners,
  before skip=\topsep,
  after skip=\topsep
}

\crefname{theorem}{Theorem}{Theorems}
\crefname{proposition}{Proposition}{Propositions}
\crefname{lemma}{Lemma}{Lemmas}
\crefname{corollary}{Corollary}{Corollaries}
\crefname{definition}{Definition}{Definitions}
\crefname{assumption}{Assumption}{Assumptions}
\crefname{remark}{Remark}{Remarks}
\crefname{problem}{Problem}{Problems}
\crefname{property}{Property}{property}

\numberwithin{equation}{section}
\numberwithin{theorem}{section}
\numberwithin{proposition}{section}
\numberwithin{definition}{section}
\numberwithin{lemma}{section}
\numberwithin{assumption}{section}
\numberwithin{remark}{section}

\usepackage[textsize=tiny]{todonotes}

\icmltitlerunning{Feature Programming for Multivariate Time Series Prediction
}

\begin{document}

\twocolumn[
\icmltitle{Feature Programming for Multivariate Time Series Prediction
}

\icmlsetsymbol{equal}{*}

\begin{icmlauthorlist}
\icmlauthor{Alex Reneau}{equal,yyy}
\icmlauthor{Jerry Yao-Chieh Hu}{equal,yyy}
\icmlauthor{Chenwei Xu}{yyy}
\icmlauthor{Weijian Li}{yyy}
\icmlauthor{Ammar Gilani}{yyy}
\icmlauthor{Han Liu}{yyy,xxx}
\end{icmlauthorlist}

\icmlaffiliation{yyy}{Department of Computer Science, University of Northwestern, Evanston, USA}
\icmlaffiliation{xxx}{Department of Statistics and Data Science, University of Northwestern, Evanston, USA}

\icmlcorrespondingauthor{Alex Reneau}{\href{mailto:alexreneau@u.northwestern.edu}{\texttt{alexreneau@u.northwestern.edu}}}
\icmlcorrespondingauthor{Jerry Yao-
Chieh Hu}{\href{mailto:jhu@u.northwestern.edu}{\texttt{jhu@u.northwestern.edu}}}
\icmlcorrespondingauthor{Chenwei Xu}{\href{mailto:cxu@u.northwestern.edu}{\texttt{cxu@u.northwestern.edu}}}
\icmlcorrespondingauthor{Weijian Li}{\href{mailto:weijianli@u.northwestern.edu}{\texttt{weijianli@u.northwestern.edu}}}
\icmlcorrespondingauthor{Ammar Gilani}{\href{mailto:ammargilani2024@u.northwestern.edu}{\texttt{ammargilani2024@u.northwestern.edu}}}
\icmlcorrespondingauthor{Han Liu}{\href{mailto:hanliu@northwestern.edu}{\texttt{hanliu@northwestern.edu}}}
\icmlkeywords{Machine Learning, ICML}

\vskip 0.3in
]

\printAffiliationsAndNotice{\icmlEqualContribution} %

\begin{abstract}
We introduce the concept of programmable feature engineering for time series modeling and propose a feature programming framework. 
This framework generates large amounts of predictive features for noisy multivariate time series while allowing users to incorporate their inductive bias with minimal effort. 
The key motivation of our framework is to view any multivariate time series as a cumulative sum of fine-grained trajectory increments, with each increment governed by a novel spin-gas dynamical Ising model. 
This fine-grained perspective motivates the development of a parsimonious set of operators that summarize multivariate time series in an abstract fashion, serving as the foundation for large-scale automated feature engineering. 
Numerically, we validate the efficacy of our method on several synthetic and real-world noisy time series datasets.
Code is available at \href{https://github.com/SirAlex900/FeatureProgramming}{\texttt{github}}.

\end{abstract}
\setlength{\parskip}{0.5em}
\setlength{\abovedisplayskip}{5pt}
\setlength{\belowdisplayskip}{5pt}
\section{Introduction}
\label{sec:intro}
We investigate the problem of automated time series feature engineering for prediction tasks in the regression setting.
A programmable feature engineering framework is proposed, named feature programming, for multivariate time series modeling. 
Our framework facilitates the automatic generation of large amounts of meaningful features from raw data.
Simultaneously, it enables the incorporation of domain knowledge through the use of feature templates, which are customizable lists of both raw and hand-crafted features provided by users.

\begin{figure*}
    \vspace{-0.05truein}
    \centering
    \includegraphics[scale=0.66]{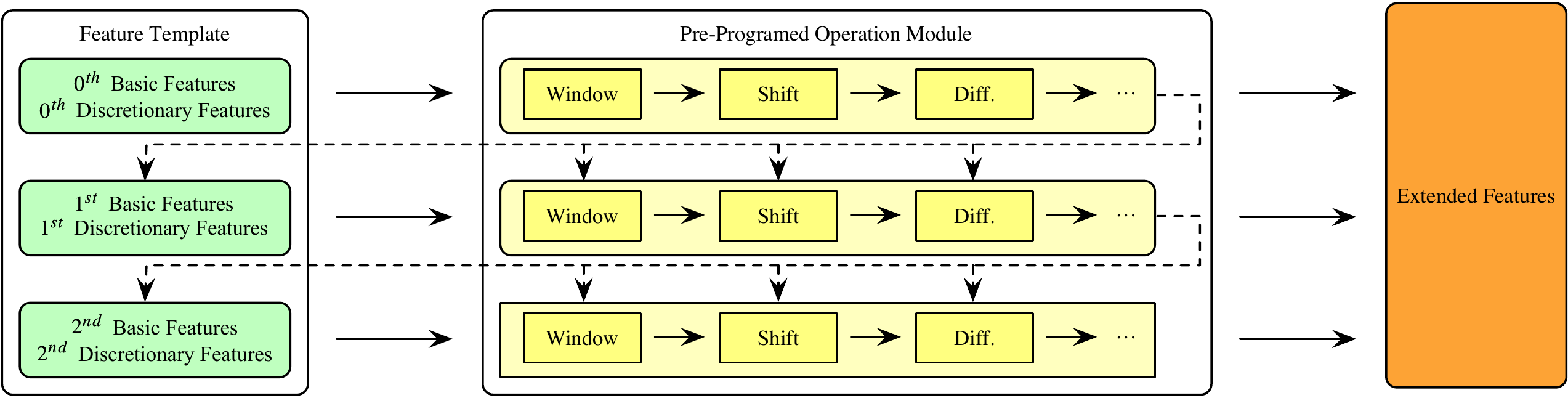}
    \vspace{-1.5em}
    \caption{\small 
    Feature Programming Pipeline. \textbf{Left:} Customizable Feature Template. \textbf{Mid:} Pre-Programmed Operation Module. \textbf{Right:} Generated Extended Features. Customization involves user-specific feature templates, operation modules, and user-designed feature flows (arrows). Solid arrows represent programmed flow within each order level, while dashed arrows indicate order-by-order feature generation flow between levels.
    See \cref{appendix:customization} for practical examples illustrating the customization procedures.
    }
    \label{fig:pipeline}
    \vspace{-0.1truein}
\end{figure*}

Our key motivation comes from  a novel dynamical Ising-like model, the spin-gas Glauber dynamics, originated from a newly debuted gas-like interaction that includes momentum and acceleration information.
By using spin-gas Glauber dynamics as the fundamental model for time series generating processes at the smallest time scale, we explore the potential of treating time series as the path-sum of infinitesimal increments generated by a series of Markovian coin tosses following the spin-gas Glauber dynamics.
From such a fine-grained perspective,  a set of operators  is motivated for extracting informative features in an abstract fashion.
We introduce the idea of the feature programming framework as a three-step pipeline for feature generation (\cref{fig:pipeline}): 
\begin{enumerate}[leftmargin=*,before=\vspace{-0.5em}, after=\vspace{-0.5em}]
\setlength\itemsep{0em}
    \item 
    Design a customized three-level feature template, which, at each level (list), includes both raw features from the data and user-specified (discretionary) features.
    \item 
    Implement a programmable operation module consisting of pre-specified operations by the user based on the proposed operator combinations.
    \item
    Generate a large number of predictive features automatically by feeding the feature template into the programmable module, following a user-specified hierarchical feature generation rule encoded in the operation module. 
\end{enumerate}
For more comprehensive details, see \cref{sec:method}, and for concrete case studies, please see \cref{appendix:customization}.

\paragraph{Contributions.} 
Our contribution is three-fold.
    \begin{itemize}[leftmargin=*,before=\vspace{-0.5em}, after=\vspace{0em}]
    \setlength\itemsep{0em}
        \item
        Methodologically, we present a novel dynamical Ising-style model for time series data generation. 
        This model motivates a parsimonious set of operators for mining valuable features from time series data. 
        Building on this, we introduce a feature programming framework, which enables the systematic generation of a large number of features by leveraging the motivated operators and a model-based hierarchical feature generation rule. 
        Additionally, this framework provides the flexibility for users to  incorporate discretionary inductive biases through feature templates.
        \item 
        Theoretically, to the best of our knowledge, our framework is the first automated feature generation method for time series equipped with a solid theoretical foundation.
        We show that appropriate selection of the interaction term in the spin-gas Glauber dynamics facilitates the derivation of a multivariate dynamical model from univariate exponential family distributions, thus enabling efficient learning and inference.
        The conditions for such a construction and its unique determination are also identified.
        \item 
        Empirically, the superiority of the generated features is demonstrated through extensive experiments on multivariate time series prediction tasks. These experiments involve various advanced models, practical scenarios, customization examples, and ablation studies, showcasing our method's adaptability, flexibility, and robust feature generation. Specifically, in our most challenging multi-horizon tasks (predicting 20 future values using a length-20 sequence, see \cref{table:multi-horizon}), our features significantly improve prediction accuracy (on average in 88+\% $R^2$ and 27+\% in Pearson correlation) across all models.
    \end{itemize}

\paragraph{Related Works.} Time series modeling has been regarded as one of the most difficult problems in machine learning for the past few decades given its temporal ordering nature and noise-sensitivity from sequential structure \cite{masini2023machine,fawaz2019deep,ozaki2012time}.
It is well-known that features play a major role in time series ML prediction problems \cite{meisenbacher2022review,cerqueira2021vest,christ2016distributed}.
While there is rich literature on data augmentation \cite{wen2020time,iwana2021empirical} and architectures \cite{tealab2018time,balkin2000automatic,sezer2020financial}, little has been done toward handling features for time series modeling and most methods prefer an end-to-end feature selection \cite{barandas2020tsfel,christ2016distributed,sun2015using,langkvist2014review,chandrashekar2014survey} or hand-crafted (pre-defined) feature design \cite{zhou2021trade,gu2020empirical,christ2018time,muralidhar2018incorporating,kakushadze2016101,christ2016distributed} without fundamental principles.

For each application domain, 
abovementioned approaches are popular and intuitive, however, they suffer from two potential issues: (i) being too purpose-specific and not transferable from one domain to another \cite{zhuang2020comprehensive,weiss2016survey,glorot2011domain,pan2010cross,pan2009survey}; 
(ii) quite some representation effort is spent on recovering the “noise” part of time series data \cite{wen2020time,iwana2021empirical,chepurko2020arda,raissi2019physics}.
To this end, our framework represents the first theoretically grounded automated feature engineering method for time series, characterized by its completeness and model-based fundamental principles.

\paragraph{Organization.}
In \cref{sec:background}, we lay out the problem setting and preliminaries.
In \cref{sec:model}, we introduce the motivating physics model.
In \cref{sec:method} we present the proposed feature programming framework.
In \cref{sec:theory}, the construction of the motivating model from univariate exponential family distributions is discussed.
In \cref{sec:exp}, experimental studies are conducted.
In \cref{sec:conclusion}, concluding discussions are provided.
More related works are discussed in \cref{appendix:related_works}.

\section{Problem Setting and Background}
\label{sec:background}

In this section, we first layout the problem setup of multivariate time series prediction and then introduce the idea we build upon: Glauber dynamics.
\subsection{Multivariate Time Series Prediction Problem}

Consider a multivariate time series dataset $\mathcal{D}$ (of size $\abs{\calD}$) with each multivariate time series being made up of $N$ correlated univariate time series,
$$
\mathcal{D}:\left\{ \((\bX_{t-T+1},\dots,\bX_t),\bY_{t+1}\)_\mu  \right\}_{\mu=1}^\abs{\calD}.
$$
$\bY_{T+1}$ is the one-step-ahead target, and
$\bX_t=\(\bx_{1,t},\dots,\bx_{N,t}\)^\sT\in \mathbb{R}^{N\times d}$ and $\bY_t=\(\by_{1,t},\dots,\by_{M,t}\)^\sT \in \mathbb{R}^{M\times d^\prime }$ are the input features and targets for each variate at each time step, respectively.

We investigate the problem of multivariate time series prediction in discrete-time in regression setting. Our goal is to predict the one-step-ahead target value, $\bY_{t+1}$, using a prediction model, $f$, that is trained on the dataset $\calD$. 
The input to the model is a sequence of past $T$ time steps, $(\bX_{t-T+1},\dots,\bX_t)$, and the output is the predicted next step $\hat{\bY}_{t+1}\coloneqq f(\bX_{t-T+1},\dots,\bX_t)$ for general sequence-to-one forecasting tasks.
Throughout this work, for a given time step $t$, we use $\bx_{i,t}\in \mathbb{R}^d$ to represent a set of covariates (features for ML models) at time step $t$, indexed by $i\in[N]$;
 we consider only scalar output for each variate  and match the number of variates by setting $d^\prime=1$ and $M=N$.

In this work, instead of investigating the effectiveness of autoregressive and cross-sectional architectures, we focus on capturing these important time series characteristics by resorting to the quality of features.
With any given architecture, we aim to improve the feature quality by engineering them to be more informative.
Namely, for each variate, in addition to using the basic (raw) features $\bx_{i,t}$ as the input, we can include engineered (extended) features $\Bar{\bx}_{i,t}\in\mathbb{R}^{\Bar{d}}$ that summarize all information from the basic features, across both time (autoregressive) and series (correlations),
for the model training.

\subsection{Dynamical Ising Model: Glauber Dynamics}

Glauber dynamics \cite{nguyen2017inverse,glauber1963time}, also known as the dynamical Ising model,
a well-studied reversible
Markov chain defined for any Markov random field originated from statistical mechanics.
It has been applied to a variety of problems, including lattice models in condensed matter physics \cite{goldenfeld1992lectures}, spin glasses \cite{janssen1976lagrangean}, neural networks \cite{mezard2009information,hinton1986learning}, and traffic \cite{pan2023Ising}. 
It has also been extended to non-equilibrium systems \cite{swendsen1987nonuniversal} and used  with mean-field theory to study complex systems \cite{montanari2022short,kadanoff2000statistical}.

To describe such a dynamical Ising model, we shall start with Ising model.
We consider an Ising model of $N$ spins as an exponential family model for binary $N$-spin data up to quadratic sufficient statistic taking the Boltzmann form
\begin{align}
    \label{eqn:boltz}
    \hspace{-2mm}
    P(\bm{\sigma}) 
    = \frac{1}{\calZ}\exp\left\{-\beta \(\sum_{e_{ij}\in \calE} J_{ij}\sigma_i\sigma_j + \sum_{v_i\in \calV} h_i \sigma_i\)\right\}
    \,,
\end{align}
where $\bm{\sigma}\coloneqq\{\sigma_1,\cdots,\sigma_N\}\in\{\pm 1\}^N$ is the configuration of $N$ binary random variables (spins) $\sigma_i\in\{\pm 1\}$  assigned to the Ising graph $\calG = (\calV,\calE)$ , $\beta\ge 0$ is the inverse temperature, and $\calZ\coloneqq \sum_{\bm{\sigma}} e^{-\beta E(\bm{\sigma})}$ is the partition function ensuring the normalization of $P(\bm{\sigma})$.
The symmetric $N\times N$ strength matrix $J\in \mathbb{R}^{N\times N}$, with zeros on the diagonal, and the external field vector $h\in \mathbb{R}^N$ encode the graphical structure of the Ising model.
For simplicity, we set the inverse temperature $\beta=1$ throughout this paper. 

The dynamics appears as we start treating \eqref{eqn:boltz} as a dynamical process and considering the time ordering of the configurations, namely with time index $\bm{\sigma}_t$.
For the purpose of this work, we introduce only the discrete-time version of Glauber dynamics.
This implies that for each time step $t$, the process begins at a potentially random initial configuration, $\sigma_t$. 
Subsequently, we assign new values to all spins independently (for each spin) for the next step in accordance with the stochastic updating rule:
\bea
\label{eqn:glauber}
P\(\sigma_{i, t+1}|\bm{\sigma}_{t}\)
=
\frac{\exp{\sigma_{i, t+1}\gamma_{i,t}}}{2\cosh{\gamma_{i,t}}},
\eea
where $\gamma_{i,t}\coloneqq \sum_j J_{ij}\sigma_j + h_i$ is the $i$th effective local field at time step $t$.
Notably, $\gamma_{i,t}$ is computed only for the $i$th spin, and the parameters of the Ising model, $(J,h)$, are time-independent.
In this work, we consider a Glauber dynamics with temporal update satisfying the following evolutionary independent assumption.
\begin{assumption}[Evolutionary Independence]
    \label{assumption:evol_indept}
    We assume the spin updates are independent events on same time slice, or equivalently, $P(\bm{\sigma}_{t+\delta t}|\bm{\sigma}_t)=\prod_{i=1}^N P(\sigma_{i,t+\delta t}| \bm{\sigma}_t)$.
\end{assumption}

\section{Spin-Gas Glauber Dynamics}
\label{sec:model}
In this section, we present the spin-gas Glauber dynamics as the motivating model for feature programming by incorporating momentum and acceleration effects into the standard Glauber dynamics to make the spin values gas-like in the effective local field.
We shall see that this modified Glauber dynamics can naturally  serve as a foundational model for multivariate time series from a fine-grained perspective.
For the ease of presentation, we reduce the feature dimension to $d=1$ by representing each covariate $\bx_t$ as a scalar $x_t \in \mathbb{R}$, and therefore $\bX_t = (x_{1,t},\dots,x_{N,t})^\sT \in \mathbb{R}^N$.
Yet, the setting of $d=1$ is primarily for illustrative purposes.
In fact, as we shall see in \cref{sec:method}, our methodology is general enough to handle $d\ge 1$.

\textbf{Fine-Graining in Time.}
Consider a multivariate time series consisting of two endpoints $\bX_t$ and $\bX_{t+\Delta t}$, separated by a time interval of $\Delta t$. 
We divide this time interval into $L$ segments by introducing a sufficiently large $L$, i.e., $\Delta t \coloneqq L \delta t$.
The multivariate time series path difference $\Delta \bX_t\coloneqq\bX_{t+\Delta t}-\bX_t$ is therefore the sum of $L$ consecutive increments for each univariate series. 
We then treat each increment as a binary outcome coin toss with $\sigma=\{\pm 1\}$ for each univariate series (or $\bm{\sigma}=\{\pm 1\}^N$ for multivariate.) 
Consequently, a random process of $L$ consecutive coin tosses generates the path difference $\Delta \bX$, and it gives the endpoint value as:
\begin{align}
\label{eqn:fine_grained_X}
\bX_{t+\Delta t}&\coloneqq
\bX_t+\Delta \bX_t
=\bX_t + c\sum_{l=1}^{L} \bm{\sigma}_{t+l\delta t} ,
\end{align}
where $c$ is a rescaling factor ensuring $\Delta \bX_t$ is bounded.

Furthermore, we assume that each coin toss with binary random variables is sampled from the conditional $\bm{\sigma}_{t+l\delta t}\sim P(\bm{\sigma}_{t+l\delta t}|\bm{\sigma}_{t+(l-1)\delta t})$, governed by the following \textit{irreversible}\footnote{By irreversible, we consider an out-of-equilibrium Glauber dynamics by generalizing $(J,h)$ to be time-dependent, see \cite{nguyen2017inverse,vazquez2017simple} and reference therein.} $N$-spin modified Glauber dynamics. 
This results in a fundamental model for time series generation from the fine-grained perspective.

\textbf{Spin-Gas Glauber Dynamics.}
We express the coin tossing process as a dynamical Ising-like system, whose time evolution is a Markov process, and hence each coin toss $\bm{\sigma}_{t+\delta t}$ (the infinitesimal increment after a $\delta t$ time interval conditioned on current time $t$) is governed by
\bea
\label{eqn:fine_grained_P}
P\(\sigma_{i, t+\delta t}|\bm{\sigma}_{t}\)
=
\frac{\exp{\sigma_{i, t+\delta t}\Gamma_{i,t}}}{2\cosh{\Gamma_{i,t}}},
\eea
where\footnote{\label{footnote:rescaling}Although the binary random variables of standard Ising model leads to discrete increments, this model of coin toss can also be used to model continuous time series by rescaling $\sigma_{i,t+\delta t}$.
That is, with a sufficiently large $L$, by taking $c\ll 1$, 
the time series $\bX_t$ becomes approximately continuous as $\Delta \bX$ becomes continuous.} $\sigma_{i,t+\delta t}=\{\pm 1\}$, and $\Gamma_{i,t}$ is a novel effective local field associated to the $i$'th spin at time $t$ defined as follows.
Let $\bp_t\coloneqq c\fdv{\bm{\sigma}_t}{t}=\frac{c(\bm{\sigma}_{t}-\bm{\sigma}_{t-\delta t})}{\delta t}$ and $ \ba_t\coloneqq c\fdv{}{t}\(\fdv{\bm{\sigma}_t}{t}\)=\frac{\bp_t-\bp_{t-\delta t}}{\delta t}$  be the discrete counterparts of momentum and accelaration on node $i$.
We have
\begin{align}
\label{eqn:local_field}
\Gamma_{i,t} \coloneqq \sum_{j} J_{ij}(t) \sigma_{j,t} + h_{i}(t) 
+
g_i\(\bp_t,\ba_t \),
\end{align}
with a newly debuted \textit{gas-like} interaction term $g$ including the effects of momentum $\bp_t$ and acceleration $ \ba_t$.
Intuitively, $\Gamma_{i,t}$ summarizes the structural contributions from all other series with the first two terms in \eqref{eqn:local_field} similar to the traditional Glauber dynamics.
In addition, with the newly introduced $g$, it includes contributions from momentum and acceleration of the series itself and all other series
at $t$, hence gas-like.
We emphasis that, the newly introduced dynamical Ising model \eqref{eqn:fine_grained_P} is the fine-grained model for the $\delta t$-infinitesimal change of multivariate time series (with stepwise change $\bm{\sigma}_t$), and serves as the fundamental dynamical model of the time series generation.

\textbf{Reverting to Continuous Time through Coarse-Graining.}
To tailor our model to handle generic time series, we must revert \eqref{eqn:fine_grained_P} back to a continuous timeframe. 
This involves a process known as coarse-graining, where we perform path-sum of $\bm{\sigma}_t$ from $\delta t$ to $L\delta t$ according to \eqref{eqn:fine_grained_X}.
Following this, we take the continuous time limit and identify the endpoint value of the trajectory, which is the cumulative sum of the random process over time starting from $\bX_t$. This approach bears similarities to the path-integral representation utilized for diffusion processes in physics, as elaborated in \cite{graham1977path,wissel1979manifolds}.

Interestingly, when each coin toss follows the Markov spin-gas Glauber dynamics \eqref{eqn:fine_grained_P}, it is possible to calculate the conditional path probability, given an initial condition $\bX_t$. 
This is achieved by first decomposing the joint distribution $P(\bm{\sigma}_{t},\ldots \bm{\sigma}_{t+L\delta t})$ as
\begin{align}
P(\bm{\sigma}_{t},\ldots \bm{\sigma}_{t+L\delta t})
=\prod_{l=1}^L 
P(\bm{\sigma}_{t+l\delta t}|\bm{\sigma}_{t+(l-1)\delta t}), 
\end{align}
and then summing out all intermediate variables ranging from $\bm{\sigma}_{t+\delta t}$ to $\bm{\sigma}_{t+(L-1)\delta t}$ such that\footnote{Recall that, since the each coin toss is Markovian, the joint distribution $P(\bm{\sigma}_t,...,\bm{\sigma}_{t+L\delta t})$ can be recursively expanded as
$\prod_{l=1}^L
P(\bm{\sigma}_{t+l\delta t}|\bm{\sigma}_{t+(l-1)\delta t})$.} 
{\footnotesize
\begin{align}
\label{eqn:discrete_path_integral}
&P\(\bX_{t+\Delta t}|{ \bX_t}\)
=P\(\bm{\sigma}_{t+L\delta t}|\bm{\sigma}_t\)\\
&=
\prod_{l=2}^L \sum_{\{\bm{\sigma}_{t+(l-1)\delta t}\}} \;
P(\bm{\sigma}_{t+l\delta t}|\bm{\sigma}_{t+(l-1)\delta t})\cdot P(\bm{\sigma}_{t+\delta t}|\bm{\sigma}_{t})\nonumber
\\
&=
\int \prod_{l=2}^L \dd \bm{\sigma}_{t+(l-1)\delta t}\;
P(\bm{\sigma}_{t+l\delta t}|\bm{\sigma}_{t+(l-1)\delta t})\cdot P(\bm{\sigma}_{t+\delta t}|\bm{\sigma}_{t})\nonumber
\\
&\propto
\int \prod_{l'=2}^L \dd \bm{\sigma}_{t+(l'-1)\delta t}
\exp\Bigg\{\sum_{l=1}^L\(\sumN \Gamma_{i,t+(l-1)\delta t} \sigma_{i,t+l\delta t}\)\Bigg\}.\nonumber
\end{align}}
To match the common notation used in physics, in the third line we abuse the notation by writing the summations over configurations as integrals, for more notational details see \cite{graham1977path}. 
Denoting $\calD\[\bm{\sigma}_\tau\] \coloneqq \prod_{l'=2}^L \dd \bm{\sigma}_{t+(l'-1)\delta t}$, the last line of \eqref{eqn:discrete_path_integral} can be understood as a well-defined discrete path integral 
{\footnotesize
\bea
P\(\bX_{t+\Delta t}|{ \bX_t}\)
\propto
\int \calD\[\bm{\sigma}_\tau\]
\exp{\sum_{l=1}^L \calL\(\bm{\sigma}_{t+(l-1)\delta t},\bm{\sigma}_{t+l\delta t} \)},
\nonumber
\eea
}
with
Lagrangian 
\begin{align}
    &\calL\(\bm{\sigma}_{t+(l-1)\delta t},\bm{\sigma}_{t+l\delta t} \)\coloneqq\nonumber\\
    &\quad\quad\sumN \Gamma_{i,t+(l-1)\delta t}\cdot \sigma_{i,t+l\delta t}-\ln\(e^{\Gamma_{i,t}}+e^{-\Gamma_{i,t}}+1\),\nonumber
\end{align}
up to a constant coefficient \cite{graham1977path}\footnote{Note that, it is a common practice to term $\calL$ as ``Lagarngian'' here since it ties the path integral formulation with the principle of least action in classical mechanics \cite{feynman2010quantum}.
More precisely, the Lagrangian $\calL$, when integrated or summed over a time period, provides us with a quantity called ``action''. 
This action makes an appearance in the exponent of  \eqref{eqn:discrete_path_integral}.
As we shall see next, this action takes a central role in the path integral formula, as the path taken by a system between two configurations is the one for which the action is minimized.}. 
Here, $\exp{\sum_{l=1}^L \calL\(\bm{\sigma}_{t+(l-1)\delta t},\bm{\sigma}_{t+l\delta t} \)}$ is a functional (the exponentiated action functional) on the space of all (discrete) paths (denoted as $\[\bm{\sigma}_\tau\]\coloneqq [\bm{\sigma}_t,\bm{\sigma}_{t+\delta t},\ldots,\bm{\sigma}_{t+L\delta t}]$), i.e., discrete functions that represent these paths.
Moreover,  $ \calD\[\bm{\sigma}_\tau\] $ is a measure of integration/summation over all possible paths $\[\bm{\sigma}_\tau\]$ connecting $\bX_t$ and $\bX_{t+\Delta t}$ in that space.
Importantly, the space of functions under consideration is defined by the details of the Lagrangian $\calL$.

When extending to continuous time limit, we reduce the time interval $\delta t \to 0$ and all infinitesimal increment size $c\bm{\sigma}_t\to 0$ (by setting $c\to 0$), and assume the multivariate $\bX_\tau$ for $\tau\in[t,t+\Delta]$ is a Markovian stochastic process with continuous sample paths.
The multivariate time series endpoint value (position) $\bX_{t+\Delta t}$ and time $t+ \Delta t$ of the system can be therefore expressed as (with the starting position $\bX_t$ and time $t$):
$\bX_t+
c\sum_{l=1}^{L} \bm{\sigma}_{t+l\delta t} \to \bX_t+\Delta\bX_{t},
$ and $t+
L\delta t \to t+\Delta t,
$
where $\Delta\bX_{t}$ and $\Delta t$ are now continuous.
The difference momentum $\fdv{\bX_t}{t}$ and acceleration $\fdv{}{t}\(\fdv{\bX_t}{t}\)$ hence become their derivative counterparts by taking large $L$ limit (with proper scaled $\sigma_{i,t+\delta t}$): $\bp_t\coloneqq\pdv{\bX_t}{t}$ and $\ba_t\coloneqq\pdv{}{t}\(\pdv{\bX_t}{t}\)$.
Consequently,  in continuous time limit, we obtain the formal expression of the continuous analog of conditional \eqref{eqn:discrete_path_integral}: 
\bea
\label{eqn:cont_path_integral}
P\(\bX_{t+\Delta t}|{ \bX_t}\)\propto \int_{\bX_t}^{\bX_{t+\Delta t}} \calD[\bX_{\tau}] \exp{-S\(\[\bX_\tau\]\)},
\eea
up to a normalization constant; where  $\calD[\bX_\tau] $ is the integration measure
in functional space of all possible paths $[\bX_\tau]$ between $\bX_t$ and $\bX_{t+\Delta t}$, $S\(\[\bX_\tau\]\)\coloneqq \int_t^{t+\Delta t} \dd\tau\; \calL(\tau)$ is the action functional. Both $\calD[\bX_\tau]$ and $S\(\[\bX_\tau\]\)$ are explicitly determined by details of the spin-gas Glauber dynamics via Lagrangian $\calL(\tau)$.  
It is worthy to note that, 
only \eqref{eqn:discrete_path_integral} is suitable for direct calculations, whereas \eqref{eqn:cont_path_integral} 
 is purely formal with more subtle use cases.
This is because  $S\(\[\bX_\tau\]\)$ cannot be trivially derived from the Riemann sum $\sum_l\calL\(\bm{\sigma}_{t+l\delta t},\bm{\sigma}_{t+(l-1)\delta t} \)$, see \cite{graham1977path,lau2007state,weber2017master} for detailed constructions and examples.

As a motivating model, 
 \eqref{eqn:discrete_path_integral} and 
the formal expression \eqref{eqn:cont_path_integral}
provide strong intuition for time series feature extraction and modeling from the data-driven perspective:
the coarse-grained model $P(\bX_{t+\Delta t}|\bX_t)$ is characterized by building blocks $\{\bm{\sigma}_{t+l\delta t}\}_{l=1}^L$, $\{\E\[\bm{\sigma}_{t+l\delta t}|\bm{\sigma}_{t+(l-1)\delta t}\]\}_{l=1}^L$, and $\{\bp_{t+l\delta t}, \ba_{t+l\delta t}\}_{l=1}^L$ for the following reasons.
Firstly, an expression like $\eqref{eqn:cont_path_integral}$ naturally suggests the most probable path between two endpoints is given by the the variation $\delta\[\int_t^{t+\Delta t} \dd\tau\; \calL(\tau)\]$, while, 
in practice, $\calL(\tau)$ is estimated from data and conditionally dependent step by step on the most recent realization according to \eqref{eqn:discrete_path_integral}.
Therefore, data from \textit{finer} timeframe(s), $\{\bm{\sigma}_{t+l\delta t}\}_{l=1}^L$
and, their \textit{smoothed} conditional expectations (conditional mean), $\{\E\[\bm{\sigma}_{t+l\delta t}|\bm{\sigma}_{t+(l-1)\delta t}\]\}_{l=1}^L$, are required to take into account both the fluctuations and the denoised most probable path in the modeling process.
Moreover, the derivative data $\{\bp_{t+l\delta t}, \ba_{t+l\delta t}\}_{l=1}^L$ from the finer timeframe(s) is also required by the definition of $\calL$.
We emphasize the building block $\{\bp_{t+l\delta t}, \ba_{t+l\delta t}\}_{l=1}^L$ is not merely a straight readout from $\calL$. Instead, as we shall see later in \cref{sec:theory}, it is 
complemented by our theoretical insights (\cref{thm:main_thm}).

We conclude this section by summarizing several advantages of modeling multivariate time series in the fine-grained perspective with the proposed dynamical Ising-like model:
(i) it accommodates both autoregressive and cross-sectional interactions;
(ii)
its strong physics intuition leads to the important characteristics for feature extraction;
(iii) as we shall see in \cref{sec:theory}, by carefully selecting the form of $g$, it has the ability to model a wide range of distributions and has closed-form multivariate densities, which enables efficient inference and learning with statistical guarantees.

\section{Methodology}
\label{sec:method}
The feature programming framework comprises three crucial components (see \cref{fig:pipeline}):
\begin{itemize}[leftmargin=*,before=\vspace{-0.5em}, after=\vspace{-0.5em}]
\setlength\itemsep{0em}
    \item 
    A set of programmable operators (Difference, Window, Shift) that form the basis for generating features; 
    \item
    A feature template that enables users to select the fundamental features they want to employ;
    \item 
    A semi-automated order-by-order feature generation rule encoded in the operation module, that automatically creates extended features within each level and between levels, adhering to an upgrade rule.    
\end{itemize}
Consequently, a feature program (which produces a set of extended features) consists of a user-specific feature template and a pre-programmed operation module. We discuss each part in the following.

\subsection{The Difference, Window and Shift Operators}
Inspired by the spin-gas Glauber dynamics, we propose three abstract operators for extracting features from time series data and discuss their operational specifics here.

\textbf{Difference Operator.}
Motivated by $\{\bp_{t+l\delta t}, \ba_{t+l\delta t}\}_{l=1}^L$, we propose the  difference operator that incorporates both continuous and finite differences and performs  series-wise subtraction between any two series. 
Operationally, we define the difference operator, 
{\small
$\mathtt{Difference[series1,series2]}$},
as the generalized derivative operation that performs first smoothing then subtracting two input series (basic features), and generates curvature-like features resembling the momentum and acceleration in physics.
With the difference operator, we characterize features (both basic and extended) into three hierarchical classes based on their \textit{order} of derivative: 0th-, 1st-, and 2nd- order features, which correspond to the generalized notion of position, momentum, and acceleration of the input raw features $\bx_{i,t}$, respectively.

\textbf{Window Operator.}
Motivated by both $\{\bm{\sigma}_{t+l\delta t}\}_{l=1}^L$ and  $\{\E\[\bm{\sigma}_{t+l\delta t}|\bm{\sigma}_{t+(l-1)\delta t}\]\}_{l=1}^L$,
we propose the window operator which summarizes the information in a fixed lookback size, denoted as $\Delta t$, from multiple resolutions using denoised summary statistics such as maximum, minimum, and mean. 
Operationally, the window operator,
{\small$\mathtt{Window[series,lookback\_size]}$},
is defined as a function that takes an input series and a 
{\small
$\mathtt{lookback\_size}$},
for each resolution, and subsequently outputs a series of summary statistics for the given lookback window. 
By applying the window operator with different lookback sizes, one can derive informative features from the time-rescaling property of the input series \cite{tallec2018can}.

\textbf{Shift Operator.}
Lastly, we propose the shift operator that can create new series with arbitrary time differences from any existing series to complement the other operators. 
The shift operator, defined as 
{\small
$\mathtt{Shift[series,\Delta\tau]}$},
allows for the input series to be shifted by any desired time difference, 
{\small
\texttt{$\Delta \tau$}},
in order to incorporate more auto-correlated information.

\subsection{Order-Upgrade Rule} 
Equipped with the aforementioned difference operator, we can generate higher-order features by applying it to lower-order features.
Then, the window and shift operators can be used to create summary-features from these higher-order features. 
For example, we can create 1-st order series by applying the difference operator to two 0-th order series, and 2-nd order series by applying the difference operator to two 1-st order series. 
Additionally, applying the window and shift operators on these higher-order series can create summary-features without changing the order of the series.

\subsection{Feature Template: Injecting Hand-Crafted Features}
The feature template is a crucial component of the feature programming framework and serves as the starting point for the feature generation process. It consists of three \textit{basic series lists}, one for each order of features (0th, 1st, 2nd) as demonstrated by the green boxes in \cref{fig:pipeline}. 
These lists are initialized as a combination of a set of basic features derived from the raw data and a discretionary design list of hand-crafted features. 
As the feature generation process progresses, these lists are updated with features from the previous order, as indicated by the dashed arrows in \cref{fig:pipeline}.
This design allows for a high degree of flexibility and customization, as the basic series lists can be fully discretionary, completely default, or a combination of the two.
Moreover, the basic series lists are fed into pre-programmed operators to generate extended features (the solid arrows in \cref{fig:pipeline}), making the setup of the basic series lists at each order a critical step in the feature generation process.

\subsection{Automate Order-by-Order Feature Generation}

With the feature template and programmable operators, we generate extended features through a semi-automated process by controlling the flow into the operation module. 
Starting with the basic series lists specified in the feature template, we feed each list into an operation module that contains the pre-programmed operators. This process is performed order-by-order, where each basic series list of an order is operated on by the corresponding operations of that order in the module. 
This results in a hierarchical feature generation that generates extended features automatically within each level and between levels, following the upgrade rule. 
Through the combinatorial manipulation of operators in the module and computation flow (represented by the arrows), our feature programming framework enables the automated and programmable generation of extended features from basic ones.
For specific examples of custom feature programs that illustrate the customization steps needed in real-world applications, please refer to \cref{appendix:customization}.

\section{Theoretical Analysis}
\label{sec:theory}

In this section, we show that, the node-conditional \eqref{eqn:fine_grained_P} uniquely specifies a joint distribution $P(\bm{\sigma}_t,\bm{\sigma}_{t+\delta t})$ under identified sufficient conditions.
This theoretical analysis further complements our physics intuitions for feature extraction in \cref{sec:model}.

\subsection{Dynamical Ising via a Temporal  Joint Graph}
Here, we formulate the joint distribution of the coin toss process, i.e. the \textit{path} probability $P(\bm{\sigma}_{t+\delta t},\bm{\sigma}_t)$ between consecutive time steps $t$ and $t+\delta 
t$, given by \eqref{eqn:fine_grained_P} as a graphical model factored according to a temporal joint graph $\calG_{t,t+\delta t}$ described below. 

Let $\calG_{t,t+\delta t}
=\(\calV_t,\calV_{t+\delta t}, \calE\)$ be the undirected temporal joint graph constituted by two subgraphs of equal cardinality, i.e. $\abs{\calV_t}=\abs{\calV_{t+\delta t}}$.
The configuration of this model, $\calG_{t,t+\delta t}$, is given as 
\begin{align}
\bX &=\(\bm{\sigma}_t,\bm{\sigma}_{t+\delta t}\)\nonumber\\
&=\(\{\sigma_{i,t}\}_{i=1}^N,\{\sigma_{i,t+\delta t}\}_{i=1}^N \)
=\(\{\sigma_q\}_{q=1}^{2N}\),
\end{align}
where $\bm{\sigma}_t$ and $\bm{\sigma}_{t+\delta t}$ are binary random vectors of length $N$.
It it important to clarify that, we use two types of node indices interchangeably in this and subsequent sections: $\sigma_{i,\tau}$ (where $\tau=t$ or $t+\delta t$) and $\sigma_{q}$. 
We use the appropriate index depending on whether we are examining the sub-graph $\calG_{\tau}$ or the joint graph $\calG_{t,t+\delta t}$.

We write the temporal joint model $\calG_{t,t+\delta t}$ to be a graphical model up to pairwise sufficient statistics and extend the sufficient statistics to include extra 1st- and 2nd- order derivatives with respect to time (at $t$).
By Assumption~\ref{assumption:evol_indept}, we have $\calE_{t+\delta t}=\emptyset$ from pairwise Markov property.
Moreover, we have $\calE_t\neq \emptyset$ and $\calE_{t,t+\delta t}\neq \emptyset$.
For simplicity, we adopt shorthand notation $\Tilde{\Phi}(\sigma_q)$ (later $\Tilde{\phi}(\sigma_q)$) for derivative-extended sufficient statistics throughout this paper.
Furthermore, we require that, for all $\calG_{t,t+\delta t}$, the time derivatives are always evaluated at $t$, not $t+\delta t$.
Explicitly, the univariate and quadratic sufficient statistics are
\bea
\label{eqn:uni_diff_SS}
    \Tilde{\Phi}\(\sigma_q\)
    = \Phi\(\sigma_q\),\quad q\in \calV_{t+\delta t},
\eea
and
\begin{align}
\label{eqn:pair_diff_SS}
    &\Tilde{\Phi}\(\sigma_q,\sigma_{q'}\)
    \\
    &=
    \begin{cases}
        \Phi\(\sigma_q,\sigma_{q'}, \fdv{\sigma_q}{t}, \fdv{\sigma_{q'}}{t}, \fdv{}{t}\(\fdv{\sigma_q}{t}\),\fdv{}{t}\(\fdv{\sigma_{q'}}{t}\)\),\;  q,q'\in \calV_{t},\\
        \Phi\(\sigma_q,\sigma_{q'}, \fdv{\sigma_q}{t},  \fdv{}{t}\(\fdv{\sigma_q}{t}\)\),\quad q\in\calV_t,q'\in \calV_{t+\delta t},\\
        \Phi\(\sigma_q,\sigma_{q'}\),\quad q,q'\in \calV_{t+\delta t}.
    \end{cases}\nonumber
\end{align}
Higher-order sufficient statistics can be easily generalized.
Consequently, the pairwise graphical model based on the graph $\calG_{t,t+\delta t}$ 
over $\bX$ takes the form
\begin{align}
\label{eqn:joint_dist_pairwise}
&P(\bm{\sigma}_t,\bm{\sigma}_{t+\delta t}) 
\propto 
\exp{w\Tilde{\Phi}(\bm{\sigma}_t,\bm{\sigma}_{t+\delta t})}
\\
&\propto 
\exp{\sum_{q\in \calV} w_q \Tilde{\Phi}(\sigma_q) + \sum_{(q,q^\prime)\in \calE} w_{qq^\prime} \Tilde{\Phi}(\sigma_q,x_{q^\prime})}.  
\nonumber
\end{align}
In the following, we aim to construct this joint distribution from its node-conditionals where the node-conditionals are specified by a univariate exponential family.

Following \cite{yang2015graphical}, we assume the node-conditional distributions of this graphical model  $\calG_{t,t+\delta t}$ 
\begin{align}
\label{eqn:node_cond_pair}
&P\(\sigma_q|\bX\) 
\\
&=\exp{\Psi_q\(\bX_{\calV\setminus q}\)\Tilde{\phi}(\sigma_q)+ B(\sigma_q) - \Bar{D}\(\bX_{\calV{\setminus q}}\)
},
\nonumber
\end{align}
follow a univariate exponential family for all $q\in \calV$,
where $\Tilde{\phi}(\sigma_q)$ is the derivative-extended \textit{univariate} sufficient statistics function for random variable $\sigma_q$.
Note that, \eqref{eqn:node_cond_pair} includes both $\sigma_q\in \calV_t$ and $\sigma_q\in \calV_{t+\delta t}$ cases, where \eqref{eqn:fine_grained_P} only corresponds to the latter.
For completeness, we further assume that, for nodes $q\in \calV_{t}$ (or equivalently $\{i,t\}\in\calV_t$), the node-conditional follows the standard Glauber dynamics \eqref{eqn:glauber} with the effective local field $\Bar{\Gamma}_{q,t}$ (or equivalently $\Bar{\Gamma}_{i,t}$):
\begin{align}
\label{eqn:completeness_cond}
P\(\sigma_q|\bX\) =
P\(\sigma_{i,t}|\bX_{\setminus \{i,t\}}\)
=
\frac{\sigma_q\cdot \Bar{\Gamma}_{i,t}}{2\cosh{\Bar{\Gamma}_{i,t}}},
\end{align}
$\forall q \notin \calV_{t+\delta t}$.
Here,
$\Bar{\Gamma}_{i,t}\coloneqq
         \sum_{p\in \calV_{\setminus\{i,t\}}} \Bar{J}_{\{i,t\},p}(t) \sigma_{p,t} + h_{i}(t) $
with $\Bar{J}(t)\in\mathbb{R}^{2N\times 2N}$ being zero-padded $J\in \mathbb{R}^{N\times N}$ and $h\in \mathbb{R}^{N}$,
where $(J,h)$ are the parameters of an Ising model factored according to subgraph $\calG_t$.

Next, we provide the theorem for the existence of a unique joint distribution whose node-conditionals are specified by the Spin-Gas Glauber dynamics \eqref{eqn:fine_grained_P} and \eqref{eqn:completeness_cond}.
    \begin{theorem}[Joint Distribution of Spin-Gas Glauber Dynamics \eqref{eqn:fine_grained_P}]
    \label{thm:main_thm}
        Suppose the joint graph $\calG_{t,t+\delta t}$ consists of cliques at most size $k$.
        The augmented Glauber dynamics \eqref{eqn:fine_grained_P}, with gas-like interaction $g$, uniquely determines a joint distribution that factors according to $\calG_{t,t+\delta t}$, if and only if, 
        \begin{enumerate}[leftmargin=14pt,label=\textbf{\Roman*:},before=\vspace{-0.5em}, after=\vspace{0em}]
            \item 
            the node-conditionals are given by
        \begin{align}
        \label{eqn:thm_node_cond}
            P\(\sigma_q| \bX\)\coloneqq
            \begin{cases}
            \frac{\sigma_q\cdot \Gamma_{q,t}}{2\cosh{\Gamma_{q,t}}},\quad \forall q \in \calV_{t+\delta t},\\
            \frac{\sigma_q\cdot \Bar{\Gamma}_{q,t}}{2\cosh{\Bar{\Gamma}_{q,t}}},\quad \forall q \in \calV_{t},
            \end{cases}
        \end{align}
        where $\Gamma_{q,t}$ is given by \eqref{eqn:fine_grained_P} and $\Bar{\Gamma}_{q,t}$ is given by \eqref{eqn:completeness_cond}.
        \item 
        for $i\in \calV_t$, $g_i$ takes the form
        \begin{align}
        &g_i\(\fdv{\bm{\sigma}_t}{t},\fdv{}{t}\(\fdv{\bm{\sigma}_t}{t}\) \)
        \nonumber\\
        &\coloneqq 
        \sum_{j\in \calV_t} G^{(1)}_{ij} (t)\fdv{\sigma_{j,t}}{t}
            + \sum_{j\in \calV_t} G^{(2)}_{ij}(t) \fdv{}{t}\(\fdv{\sigma_{j,t}}{t}\)
        \nonumber\\
        \label{eqn:g_tensor_factor}
        &+ \sum_{m,n\in \calV_t} G^{(1)}_{imn} (t)\fdv{\sigma_{m,t}}{t}\fdv{\sigma_{n,t}}{t}
        \\
        &+ \sum_{m,n\in \calV_t} G^{(2)}_{imn}(t) \fdv{}{t}\(\fdv{\sigma_{m,t}}{t}\)
        \fdv{}{t}\(\fdv{\sigma_{n,t}}{t}\)+\cdots,
        \nonumber
        \end{align}
        which is up to $(k-1)$-th order product with coupling constants $\{G\}$ for 1st- and 2nd-order time derivatives and their cross terms.
        For $i\in \calV_{t+\delta t}$, $g_i=0$.
        \end{enumerate}
    \end{theorem}
\begin{proof}
A detailed proof is shown in the \cref{proof:main_thm}.
\end{proof}
\vspace{-0.1truein}
Here, we state the theorem in the general form where $\calG_{t,t+\delta t}$ consists up to $k$-cliques, and then, by setting $k=2$, it reduces to pairwise joint distribution \eqref{eqn:joint_dist_pairwise} of our interest.
\begin{corollary}
\label{corollary:A}
The corresponding joint distribution in \cref{thm:main_thm} is in the form
{\footnotesize
        \begin{align}
        &P(\bX)=\nonumber\\
        &
        \exp\bigg\{
        \sum_{q\in\calV} \Tilde{w}_q \Tilde{\phi}(\sigma_q) + \sum_{q\in\calV} \sum_{p\in\calN(q)} \Tilde{w}_{qp} \Tilde{\phi}(\sigma_q)\Tilde{\phi}(\sigma_p)+\cdots
        \nonumber\\
        &+\sum_{q\in\calV}\sum_{p_1,\cdots,p_k\in\calN(q)} \Tilde{w}_{q,p_1,\cdots,p_k} \Tilde{\phi}(\sigma_q)\prod_{j=2}^k\Tilde{\phi}(\sigma_{p_k})
        -A(\{\Tilde{w}\}) 
        \bigg\},\nonumber
        \end{align}
        }
        with derivative-extended sufficient statistics $\Tilde{\phi}(\sigma_q)\coloneqq \(\sigma_{\sigma_q},\fdv{\sigma_{\sigma_q}}{t},\fdv{}{t}\(\fdv{\sigma_{\sigma_q}}{t}\)\)$ being a 3-vector for $q\in\calV_t$, $\Tilde{\phi}(\sigma_q)=\sigma_q$ being a zero-padded 3-vector, for $q\in\calV_{t+\delta t}$ and $\{\Tilde{w}\}$ being contants 1-1 corresponding to $\{G\}$.
\end{corollary}
\begin{corollary}
\label{corollary:B}
    For $k=2$, we have pairwise model
    {\footnotesize
    \begin{align}
    &P(\bX)=\nonumber\\
    &\exp{\sum_{q\in\calV} h_q(t)\sigma_q
    +\sum_{q\in\calV} \sum_{p\in\calN(q)}  \Tilde{w}_{qp} \Tilde{\phi}(\sigma_q)\Tilde{\phi}(\sigma_p)
    -A(\{\Tilde{w}\}) },\nonumber
    \end{align}
    }
    where, for $q\in\calV_t$,
$
         \half\sum_{p\in\calN(q)} \Tilde{w}_{qp} \Tilde{\phi}(\sigma_p)=
         \sum_{p\in\calN(q)} J_{qp}(t) \sigma_{p,t}
        + \sum_{p\in\calN(q)} G^{(1)}_{qp} (t)\fdv{\sigma_{p,t}}{t}
        + \sum_{p\in\calN(q)} G^{(2)}_{qp}(t) \fdv{}{t}\(\fdv{\sigma_{p,t}}{t}\)
$
     and, for $q\in\calV_{t+\delta t}$,
         $\sum_{p\in\calN(q)} \Tilde{w}_p \Tilde{\phi}(\sigma_p)=
         \sum_{p\in\calN(q)} \Bar{J}_{qp}(t) \sigma_{p,t}$.
\end{corollary}
\begin{proof}
    Corollary~\ref{corollary:A} and Corollary~\ref{corollary:B} are direct consequences of \cref{thm:main_thm}.
\end{proof} 
 
\begin{table*}[t]
    \centering
    \vspace{-1em}
    \caption{Comparison of Basic and Extended Feature Accuracy.}
    \vspace*{0.05truein}
    \begin{tabular}{clcccccc}
    \toprule
    \toprule
        \multirow{2}{*}{Metric}& \multirow{2}{*}{Dataset} & \multicolumn{2}{c}{MLP} & \multicolumn{2}{c}{CNN} &  \multicolumn{2}{c}{LSTM} \\
        \cmidrule{3-8}
        & & Basic & Extended & Basic & Extended & Basic & Extended\\ 
        \midrule
        \multirow{4}{*}{$R^2$ Score \%} & {\small Synthetic} &  97.71 {\tiny $\pm$ 0.00} & \textbf{99.18 {\tiny $\pm$ 0.00}} & 97.72 {\tiny $\pm$ 0.00} & \textbf{99.16 {\tiny $\pm$ 0.00}} &  97.71 {\tiny $\pm$ 0.03} & \textbf{99.01 {\tiny $\pm$ 0.04}} \\
        & {\small Taxi} &  73.21 {\tiny $\pm$ 0.00} & \textbf{77.50 {\tiny $\pm$ 0.01}} & 73.17 {\tiny $\pm$ 0.00} & \textbf{79.02 {\tiny $\pm$ 0.06}} &  73.19 {\tiny $\pm$ 0.01} & \textbf{76.62 {\tiny $\pm$ 0.04}} \\
        & {\small Electricity} &  97.47 {\tiny $\pm$ 0.00} & \textbf{98.97 {\tiny $\pm$ 0.00}} & 97.47 {\tiny $\pm$ 0.00} & \textbf{99.09 {\tiny $\pm$ 0.01}} &  94.83 {\tiny $\pm$ 0.00} & \textbf{95.43 {\tiny $\pm$ 0.00}} \\
        & {\small Traffic} &  75.04 {\tiny $\pm$ 0.00} & \textbf{87.41 {\tiny $\pm$ 0.00}} & 75.04 {\tiny $\pm$ 0.00} & \textbf{86.45 {\tiny $\pm$ 0.01}} &  74.66 {\tiny $\pm$ 0.00} & \textbf{83.12 {\tiny $\pm$ 0.00}} \\
        \midrule
        \multirow{4}{*}{\makecell{Pearson \\ Correlation} \%} &
         {\small Synthetic} &  98.86 {\tiny $\pm$ 0.00} & \textbf{99.60 {\tiny $\pm$ 0.00}} & 98.86 {\tiny $\pm$ 0.00} & \textbf{99.62 {\tiny $\pm$ 0.00}} &  98.86 {\tiny $\pm$ 0.01} & \textbf{99.56 {\tiny $\pm$ 0.01}} \\
        & {\small Taxi} &  85.59 {\tiny $\pm$ 0.00} & \textbf{88.76 {\tiny $\pm$ 0.00}} & 85.57 {\tiny $\pm$ 0.00} & \textbf{88.95 {\tiny $\pm$ 0.03}} &  85.58 {\tiny $\pm$ 0.01} & \textbf{88.46 {\tiny $\pm$ 0.00}} \\
        & {\small Electricity} &  98.73 {\tiny $\pm$ 0.00} & \textbf{99.49 {\tiny $\pm$ 0.00}} & 98.73 {\tiny $\pm$ 0.00} & \textbf{99.54 {\tiny $\pm$ 0.00}} &  \textbf{98.04 {\tiny $\pm$ 0.00}} & 97.70 {\tiny $\pm$ 0.00} \\
        & {\small Traffic} &  86.64 {\tiny $\pm$ 0.00} & \textbf{93.51 {\tiny $\pm$ 0.00}} & 86.64 {\tiny $\pm$ 0.00} & \textbf{93.02 {\tiny $\pm$ 0.00}} &  86.42 {\tiny $\pm$ 0.00} & \textbf{91.40 {\tiny $\pm$ 0.00}} \\
        \bottomrule
        \bottomrule
    \end{tabular}
    \label{table:results}
\end{table*}

Following are two conclusions of our theoretical study that provide practical guidance for our methodology.
\begin{enumerate}[leftmargin=14pt,label=(\roman*),before=\vspace{-0.5em}, after=\vspace{-0.5em}]
\setlength\itemsep{0em}
    \item 
    Although Theorem \ref{thm:main_thm} does not delve into the analysis of statistical estimation for such a model, it does offer methodological guidance that mirrors the neighborhood-estimation\footnote{Estimate the neighborhood of each node separately, and then stitch then together to form the global graph estimation.} approach used for pairwise graphical models in \cite{yang2015graphical}. 
    This suggests that the ground-truth random process can be uniquely represented by derivative-extended sufficient statistics. 
    This insight is crucial for feature engineering, especially in cases where the target model is unknown.

    \item
    In the same vein, \cref{corollary:B}, in conjunction with \cite{yang2015graphical}, proposes an amenable form of $g$ to enhance learning and inference. 
    This aligns with our physics intuitions from \cref{sec:model}, which regard $\{\bp,\ba\}$ as fundamental characteristics for feature extraction.
\end{enumerate}

\section{Experimental Studies}
\label{sec:exp}
In this section, we demonstrate the validity of our feature generation method by testing it on a synthetic dataset and three real-world datasets, with three distinct neural network architectures (MLP, CNN, LSTM) representing three aspects of deep learning architecture design.
To evaluate the generated features, we use a predictive model, $F$, that only takes in univariate $\bx_{i,t}$ as input and predicts the one-step-ahead target $\by_{i,t+1}$, i.e. $F(\bx_{i,t})=\hat{\by}_{i,t+1}$, $\forall i,t$. 
By doing this, the model does not consider the correlations between univariate series or the autocorrelation beyond a single time step. This means that the advantages of using autoregressive and graphical (cross-sectional or convolutional) architectures are limited. 
To effectively capture important time series characteristics, we need input features that can summarize all information from the basic/raw features across both time (autoregressive) and univariates (correlations) for model training.
Additional experiments including more deep learning models for both one-step-ahead and multi-horizon prediction tasks can be found in \cref{appendix:extra_exp}.

\subsection{Experimental Setup}
\label{sec:setting}
\textbf{Data.}
The data utilized in our experiments consists of a synthetic dataset constructed to adhere to the assumptions of our method, as well as an electricity dataset, a traffic dataset, and a taxi dataset.
Each of these datasets is partitioned in an 80/20 ratio to derive our training data (known as in-sample data) and testing data (referred to as out-of-sample data).
Further details regarding the datasets can be found in \cref{appendix:data}.

\textbf{Feature Programming.}
We set the feature template to \textit{default}\footnote{It is important to notice that one major advantage of our feature programming framework is that it is hyperparameter-free. The generated features are solely controlled by user-designed feature program (feature template and operation module).} (without additional hand-crafted features).
We set the order-to-order generation rule to pass \textit{all} extended features from the preceding order along with the basic series list of the current order to the programmed module.
For each dataset, we generate the 0th order, 1st order, and 2nd order features.
These features are then combined and treated as extended features.
We trained three deep learning models, MLP, CNN, and LSTM, using input features consisting of
(i) only the basic features; and (ii) a concatenation of basic features and extended features.

\textbf{Benchmark and Evaluation Metrics.}
We train each of the models described above using (i) as the benchmark. 
Namely, we compare the performance of models trained on the basic features vs. models trained on both the basic and extended features. 
For our evaluation metrics, we use out-of-sample $R^2$-score and Pearson correlation between the predicted and true value (one-step-ahead).
For each dataset and all models, we repeat the multivariate time series prediction task 10 times with basic and extended features. 
We defer implementation details to \cref{appendix:implementation}.

\subsection{Evaluation of Accuracy}
In \cref{table:results}, our results demonstrate that the proposed methods can improve model accuracy significantly with high quality features.
Comparing across all datasets (except for electricity-LSTM), models that were trained incorporating both basic and extended features consistently outperformed those trained solely on basic features.
The introduction of extended features led to an augmentation in both the $R^2$ Score and the Pearson correlation. The magnitude of this improvement ranged between 1.3\% and 5.85\%, depending on the specific dataset and model utilized.

Notably, even with the most challenging dataset (the \textbf{taxi} dataset), the inclusion of extended features facilitated an improvement of approximately 4.3/5.9/3.4\% in the $R^2$ score for MLP/CNN/LSTM, and around 2.3/3.4/3.0\% in the Pearson correlations for MLP/CNN/LSTM.
The results are summarized in \cref{table:results}.

\subsection{Computational Time Comparison}

As seen in \cref{table:comp_time}, the feature programming framework demonstrates high efficiency. The computational time of feature generation is negligible compared to the time spent on training the downstream models, even when considering the MLP model, which is the simplest and fastest-to-train prediction model in our experiments. 
\begin{table}[h]
\vspace{-0.08truein}
    \centering
    \caption{Computational Time Comparison for Feature Generation and Model Training.
    This table presents the computational time required for generating extended features based on the default setting described in \cref{sec:setting}, alongside the time required for downstream model training. 
    We report the computational time of  simplest (and fastest) model in our experiments, i.e. MLP.
    The feature programming framework demonstrates high efficiency, with the feature generation time being negligible in comparison to the time spent on training the downstream models.
    }
    \vspace*{0.05truein}
\resizebox{ 0.5\textwidth}{!}{    
    \begin{tabular}{cccccccc}
    \toprule
    \toprule
         & \multicolumn{1}{c}{Taxi} 
         & \multicolumn{1}{c}{Electric} 
         & \multicolumn{1}{c}{Traffic} 
         &  \multicolumn{1}{c}{MLP-Synthetic}
         &  \multicolumn{1}{c}{MLP-Taxi}
         &  \multicolumn{1}{c}{MLP-Electric}
         &  \multicolumn{1}{c}{MLP-Traffic}
         \\ 
        \midrule
        \makecell{\rotatebox{0}{\small Time (s)}} 
        & 24.34  
        & 82.71 
        & 218.23
        & 4272.36
        & 5461.75
        & 16833.59
        & 23883.94
        \\ \bottomrule\bottomrule
    \end{tabular}
    \label{table:comp_time}
}
    \vspace*{-0.1truein}
\end{table} 

\subsection{Additional Experiments}

Further experiments on the quality of extended features
using a variety of common time series prediction models and ablation studies 
can be found in \cref{appendix:extra_exp}, including
\begin{itemize}[leftmargin=*,before=\vspace{-0.5em}, after=\vspace{-0.5em}]
\setlength\itemsep{0em}
    \item \textbf{\cref{appendix:many_models}:}
    Six commonly-used time series forecasting models (Transformer \cite{vaswani2017attention}, XGBM \cite{chen2015xgboost}, LightGBM \cite{ke2017lightgbm}, TFT \cite{lim2021temporal}, TCN \cite{chen2020probabilistic}, NBEATS \cite{oreshkin2019n}), with an emphasis on standard one-step-ahead predictions. In addition, we also incorporate multi-horizon prediction task experiments to showcase practical applicability.
    \item \textbf{\cref{appendix:customization}:}
    Ablation studies that utilize three illustrative examples to demonstrate the customization process that may be necessary in real-world scenarios. These examples illustrate how to adjust the feature program to resemble known hand-crafted features, corresponding to practical situations where we have some understanding of the types of features that might be particularly useful for a specific task, domain, or application. 
    This approach allows for a clear understanding of how to customize the feature program and how the operators influence the generated features.
    We evaluate feature quality using the same metrics across different feature programming settings.
\end{itemize}
Through our experiments, the performance enhancements provided by our extended features are illustrated. 
In particular, in the multi-horizon prediction tasks (\cref{table:multi-horizon}), our results show that when predicting the next 20 values using a sequence of length 20 (arguably the most challenging task when relying solely on basic features), the features generated by our approach have exhibited substantial improvements (on average 88+\% in $R^2$ and 27+\% in Pearson correlation metrics) in prediction accuracy across all models. 
These experiments showcase the adaptability, flexibility, and feature generation capabilities of our method.

\section{Conclusion}
\label{sec:conclusion}We propose a programmable automated feature engineering approach for multivariate time series modeling, called feature programming, inspired by a novel Ising-like dynamical model. 
Theoretically, our model-based approach draws practical guidance from constructing multivariate graphical models using univariate exponential family, aligning with insights from the physics model. 
Empirically, the generated features effectively improve noisy multivariate time series prediction in various settings. 
Yet, there is a noticeable limitation:
the flexibility of our method comes with the trade-off of not including any feature selection or pruning mechanism beyond user-specific programs.
For future investigations, we plan to integrate feature selection mechanisms into feature programming and explore the joint temporal graph in the context of graphical diffusion \cite{kondor2002diffusion}.

\section*{Acknowledgments}
JH would like to thank to Donglin Yang and Andrew Chen for enlightening discussions, and Jiayi Wang for invaluable support in facilitating experimental deployments.
The authors would like to thank the anonymous reviewers for constructive comments and the program chairs for their assistance in preparing the camera-ready version.
JH was partially supported by the Walter P. Murphy Fellowship.
HL was partially supported by NIH R01LM1372201, NSF CAREER1841569 and a NSF TRIPODS1740735.
The content is solely the responsibility of the authors and does not necessarily represent the official
views of the funding agencies.

\clearpage
\bibliography{refs}

\begin{thebibliography}{69}
\providecommand{\natexlab}[1]{#1}
\providecommand{\url}[1]{\texttt{#1}}
\expandafter\ifx\csname urlstyle\endcsname\relax
  \providecommand{\doi}[1]{doi: #1}\else
  \providecommand{\doi}{doi: \begingroup \urlstyle{rm}\Url}\fi

\bibitem[Balkin \& Ord(2000)Balkin and Ord]{balkin2000automatic}
Balkin, S.~D. and Ord, J.~K.
\newblock Automatic neural network modeling for univariate time series.
\newblock \emph{International Journal of Forecasting}, 16\penalty0
  (4):\penalty0 509--515, 2000.

\bibitem[Barandas et~al.(2020)Barandas, Folgado, Fernandes, Santos, Abreu,
  Bota, Liu, Schultz, and Gamboa]{barandas2020tsfel}
Barandas, M., Folgado, D., Fernandes, L., Santos, S., Abreu, M., Bota, P., Liu,
  H., Schultz, T., and Gamboa, H.
\newblock Tsfel: Time series feature extraction library.
\newblock \emph{SoftwareX}, 11:\penalty0 100456, 2020.

\bibitem[Bengio et~al.(2013)Bengio, Courville, and
  Vincent]{bengio2013representation}
Bengio, Y., Courville, A., and Vincent, P.
\newblock Representation learning: A review and new perspectives.
\newblock \emph{IEEE transactions on pattern analysis and machine
  intelligence}, 35\penalty0 (8):\penalty0 1798--1828, 2013.

\bibitem[Besag(1974)]{besag1974spatial}
Besag, J.
\newblock Spatial interaction and the statistical analysis of lattice systems.
\newblock \emph{Journal of the Royal Statistical Society: Series B
  (Methodological)}, 36\penalty0 (2):\penalty0 192--225, 1974.

\bibitem[Bousmalis et~al.(2016)Bousmalis, Trigeorgis, Silberman, Krishnan, and
  Erhan]{bousmalis2016domain}
Bousmalis, K., Trigeorgis, G., Silberman, N., Krishnan, D., and Erhan, D.
\newblock Domain separation networks.
\newblock \emph{Advances in neural information processing systems}, 29, 2016.

\bibitem[Cerqueira et~al.(2021)Cerqueira, Moniz, and Soares]{cerqueira2021vest}
Cerqueira, V., Moniz, N., and Soares, C.
\newblock Vest: Automatic feature engineering for forecasting.
\newblock \emph{Machine Learning}, pp.\  1--23, 2021.

\bibitem[Chandrashekar \& Sahin(2014)Chandrashekar and
  Sahin]{chandrashekar2014survey}
Chandrashekar, G. and Sahin, F.
\newblock A survey on feature selection methods.
\newblock \emph{Computers \& Electrical Engineering}, 40\penalty0 (1):\penalty0
  16--28, 2014.

\bibitem[Chen et~al.(2015)Chen, He, Benesty, Khotilovich, Tang, Cho, Chen,
  Mitchell, Cano, Zhou, et~al.]{chen2015xgboost}
Chen, T., He, T., Benesty, M., Khotilovich, V., Tang, Y., Cho, H., Chen, K.,
  Mitchell, R., Cano, I., Zhou, T., et~al.
\newblock Xgboost: extreme gradient boosting.
\newblock \emph{R package version 0.4-2}, 1\penalty0 (4):\penalty0 1--4, 2015.

\bibitem[Chen et~al.(2020)Chen, Kang, Chen, and Wang]{chen2020probabilistic}
Chen, Y., Kang, Y., Chen, Y., and Wang, Z.
\newblock Probabilistic forecasting with temporal convolutional neural network.
\newblock \emph{Neurocomputing}, 399:\penalty0 491--501, 2020.

\bibitem[Chepurko et~al.()Chepurko, Marcus, Zgraggen, Fernandez, Kraska, and
  Karger]{chepurko2020arda}
Chepurko, N., Marcus, R., Zgraggen, E., Fernandez, R.~C., Kraska, T., and
  Karger, D.
\newblock Arda: Automatic relational data augmentation for machine learning.
\newblock \emph{Proceedings of the VLDB Endowment}, 13\penalty0 (9).

\bibitem[Christ et~al.(2016)Christ, Kempa-Liehr, and
  Feindt]{christ2016distributed}
Christ, M., Kempa-Liehr, A.~W., and Feindt, M.
\newblock Distributed and parallel time series feature extraction for
  industrial big data applications.
\newblock \emph{arXiv preprint arXiv:1610.07717}, 2016.

\bibitem[Christ et~al.(2018)Christ, Braun, Neuffer, and
  Kempa-Liehr]{christ2018time}
Christ, M., Braun, N., Neuffer, J., and Kempa-Liehr, A.~W.
\newblock Time series feature extraction on basis of scalable hypothesis tests
  (tsfresh--a python package).
\newblock \emph{Neurocomputing}, 307:\penalty0 72--77, 2018.

\bibitem[Clifford(1990)]{clifford1990markov}
Clifford, P.
\newblock Markov random fields in statistics.
\newblock \emph{Disorder in physical systems: A volume in honour of John M.
  Hammersley}, pp.\  19--32, 1990.

\bibitem[De~Brabandere et~al.(2022)De~Brabandere, Op~De~Be{\'e}ck, Hendrickx,
  Meert, and Davis]{de2022tsfuse}
De~Brabandere, A., Op~De~Be{\'e}ck, T., Hendrickx, K., Meert, W., and Davis, J.
\newblock Tsfuse: Automated feature construction for multiple time series data.
\newblock \emph{Machine Learning}, pp.\  1--56, 2022.

\bibitem[Doersch et~al.(2015)Doersch, Gupta, and
  Efros]{doersch2015unsupervised}
Doersch, C., Gupta, A., and Efros, A.~A.
\newblock Unsupervised visual representation learning by context prediction.
\newblock In \emph{Proceedings of the International Conference on Computer
  Vision}, pp.\  1422--1430, 2015.

\bibitem[Fawaz et~al.(2019)Fawaz, Forestier, Weber, Idoumghar, and
  Muller]{fawaz2019deep}
Fawaz, H.~I., Forestier, G., Weber, J., Idoumghar, L., and Muller, P.-A.
\newblock Deep learning for time series classification: a review.
\newblock \emph{Data mining and knowledge discovery}, 33\penalty0 (4):\penalty0
  917--963, 2019.

\bibitem[Feynman et~al.(2010)Feynman, Hibbs, and Styer]{feynman2010quantum}
Feynman, R.~P., Hibbs, A.~R., and Styer, D.~F.
\newblock \emph{Quantum mechanics and path integrals}.
\newblock Courier Corporation, 2010.

\bibitem[Glauber(1963)]{glauber1963time}
Glauber, R.~J.
\newblock Time-dependent statistics of the ising model.
\newblock \emph{Journal of mathematical physics}, 4\penalty0 (2):\penalty0
  294--307, 1963.

\bibitem[Glorot et~al.(2011)Glorot, Bordes, and Bengio]{glorot2011domain}
Glorot, X., Bordes, A., and Bengio, Y.
\newblock Domain adaptation for large-scale sentiment classification: A deep
  learning approach.
\newblock In \emph{Proceedings of the 28th international conference on machine
  learning (ICML-11)}, pp.\  513--520, 2011.

\bibitem[Goldenfeld(1992)]{goldenfeld1992lectures}
Goldenfeld, N.
\newblock \emph{Lectures on phase transitions and the renormalization group}.
\newblock Reading, MA: Addison-Wesley, 1992.

\bibitem[Goodfellow et~al.(2014)Goodfellow, Pouget-Abadie, Mirza, Xu,
  Warde-Farley, Ozair, Courville, and Bengio]{goodfellow2014generative}
Goodfellow, I., Pouget-Abadie, J., Mirza, M., Xu, B., Warde-Farley, D., Ozair,
  S., Courville, A., and Bengio, Y.
\newblock Generative adversarial nets.
\newblock \emph{Advances in neural information processing systems}, pp.\
  2672--2680, 2014.

\bibitem[Graham(1977)]{graham1977path}
Graham, R.
\newblock Path integral formulation of general diffusion processes.
\newblock \emph{Zeitschrift f{\"u}r Physik B Condensed Matter}, 26\penalty0
  (3):\penalty0 281--290, 1977.

\bibitem[Gu et~al.(2020)Gu, Kelly, and Xiu]{gu2020empirical}
Gu, S., Kelly, B., and Xiu, D.
\newblock Empirical asset pricing via machine learning.
\newblock \emph{The Review of Financial Studies}, 33\penalty0 (5):\penalty0
  2223--2273, 2020.

\bibitem[Guyon \& Elisseeff(2003)Guyon and Elisseeff]{guyon2003introduction}
Guyon, I. and Elisseeff, A.
\newblock An introduction to variable and feature selection.
\newblock \emph{Journal of machine learning research}, 3\penalty0
  (Mar):\penalty0 1157--1182, 2003.

\bibitem[Hinton et~al.(2012)Hinton, Deng, Yu, Dahl, Mohamed, Jaitly, Senior,
  Vanhoucke, Nguyen, Sainath, et~al.]{hinton2012deep}
Hinton, G., Deng, L., Yu, D., Dahl, G.~E., Mohamed, A.-r., Jaitly, N., Senior,
  A., Vanhoucke, V., Nguyen, P., Sainath, T.~N., et~al.
\newblock Deep neural networks for acoustic modeling in speech recognition: The
  shared views of four research groups.
\newblock \emph{IEEE Signal processing magazine}, 29\penalty0 (6):\penalty0
  82--97, 2012.

\bibitem[Hinton et~al.(1986)Hinton, Sejnowski, et~al.]{hinton1986learning}
Hinton, G.~E., Sejnowski, T.~J., et~al.
\newblock Learning and relearning in boltzmann machines.
\newblock \emph{Parallel distributed processing: Explorations in the
  microstructure of cognition}, 1\penalty0 (282-317):\penalty0 2, 1986.

\bibitem[Iwana \& Uchida(2021)Iwana and Uchida]{iwana2021empirical}
Iwana, B.~K. and Uchida, S.
\newblock An empirical survey of data augmentation for time series
  classification with neural networks.
\newblock \emph{Plos one}, 16\penalty0 (7):\penalty0 e0254841, 2021.

\bibitem[Janssen(1976)]{janssen1976lagrangean}
Janssen, H.
\newblock On a lagrangean for classical field dynamics and renormalization
  group calculations of dynamical critical properties.
\newblock \emph{Zeitschrift f{\"u}r Physik B Condensed Matter}, 23\penalty0
  (2):\penalty0 377--380, 1976.

\bibitem[Jiang et~al.(2021)Jiang, Luo, Wei, Abualigah, and
  Zhou]{jiang2021efficient}
Jiang, Y., Luo, Q., Wei, Y., Abualigah, L., and Zhou, Y.
\newblock An efficient binary gradient-based optimizer for feature selection.
\newblock \emph{Math. Biosci. Eng}, 18\penalty0 (4):\penalty0 3813--3854, 2021.

\bibitem[Kadanoff(2000)]{kadanoff2000statistical}
Kadanoff, L.~P.
\newblock \emph{Statistical physics: Statics, dynamics and renormalization}.
\newblock World Scientific, 2000.

\bibitem[Kakushadze(2016)]{kakushadze2016101}
Kakushadze, Z.
\newblock 101 formulaic alphas.
\newblock \emph{Wilmott}, 2016\penalty0 (84):\penalty0 72--81, 2016.

\bibitem[Kaul et~al.(2017)Kaul, Maheshwary, and Pudi]{kaul2017autolearn}
Kaul, A., Maheshwary, S., and Pudi, V.
\newblock Autolearn—automated feature generation and selection.
\newblock In \emph{2017 IEEE International Conference on data mining (ICDM)},
  pp.\  217--226. IEEE, 2017.

\bibitem[Ke et~al.(2017)Ke, Meng, Finley, Wang, Chen, Ma, Ye, and
  Liu]{ke2017lightgbm}
Ke, G., Meng, Q., Finley, T., Wang, T., Chen, W., Ma, W., Ye, Q., and Liu,
  T.-Y.
\newblock Lightgbm: A highly efficient gradient boosting decision tree.
\newblock \emph{Advances in neural information processing systems}, 30, 2017.

\bibitem[Khurana et~al.(2016)Khurana, Turaga, Samulowitz, and
  Parthasrathy]{khurana2016cognito}
Khurana, U., Turaga, D., Samulowitz, H., and Parthasrathy, S.
\newblock Cognito: Automated feature engineering for supervised learning.
\newblock In \emph{2016 IEEE 16th International Conference on Data Mining
  Workshops (ICDMW)}, pp.\  1304--1307. IEEE, 2016.

\bibitem[Kondor \& Lafferty(2002)Kondor and Lafferty]{kondor2002diffusion}
Kondor, R.~I. and Lafferty, J.
\newblock Diffusion kernels on graphs and other discrete structures.
\newblock In \emph{Proceedings of the 19th international conference on machine
  learning}, volume 2002, pp.\  315--322, 2002.

\bibitem[L{\"a}ngkvist et~al.(2014)L{\"a}ngkvist, Karlsson, and
  Loutfi]{langkvist2014review}
L{\"a}ngkvist, M., Karlsson, L., and Loutfi, A.
\newblock A review of unsupervised feature learning and deep learning for
  time-series modeling.
\newblock \emph{Pattern Recognition Letters}, 42:\penalty0 11--24, 2014.

\bibitem[Lau \& Lubensky(2007)Lau and Lubensky]{lau2007state}
Lau, A.~W. and Lubensky, T.~C.
\newblock State-dependent diffusion: Thermodynamic consistency and its path
  integral formulation.
\newblock \emph{Physical Review E}, 76\penalty0 (1):\penalty0 011123, 2007.

\bibitem[Le \& Mikolov(2014)Le and Mikolov]{le2014distributed}
Le, Q.~V. and Mikolov, T.
\newblock Distributed representations of sentences and documents.
\newblock In \emph{International Conference on Machine Learning}, pp.\
  1188--1196, 2014.

\bibitem[Lim et~al.(2021)Lim, Ar{\i}k, Loeff, and Pfister]{lim2021temporal}
Lim, B., Ar{\i}k, S.~{\"O}., Loeff, N., and Pfister, T.
\newblock Temporal fusion transformers for interpretable multi-horizon time
  series forecasting.
\newblock \emph{International Journal of Forecasting}, 37\penalty0
  (4):\penalty0 1748--1764, 2021.

\bibitem[Masini et~al.(2023)Masini, Medeiros, and Mendes]{masini2023machine}
Masini, R.~P., Medeiros, M.~C., and Mendes, E.~F.
\newblock Machine learning advances for time series forecasting.
\newblock \emph{Journal of economic surveys}, 37\penalty0 (1):\penalty0
  76--111, 2023.

\bibitem[Meisenbacher et~al.(2022)Meisenbacher, Turowski, Phipps, R{\"a}tz,
  M{\"u}ller, Hagenmeyer, and Mikut]{meisenbacher2022review}
Meisenbacher, S., Turowski, M., Phipps, K., R{\"a}tz, M., M{\"u}ller, D.,
  Hagenmeyer, V., and Mikut, R.
\newblock Review of automated time series forecasting pipelines.
\newblock \emph{Wiley Interdisciplinary Reviews: Data Mining and Knowledge
  Discovery}, 12\penalty0 (6):\penalty0 e1475, 2022.

\bibitem[Mezard \& Montanari(2009)Mezard and Montanari]{mezard2009information}
Mezard, M. and Montanari, A.
\newblock \emph{Information, physics, and computation}.
\newblock Oxford University Press, 2009.

\bibitem[Miikkulainen et~al.(2019)Miikkulainen, Liang, Meyerson, Rawal, Fink,
  Francon, Raju, Shahrzad, Navruzyan, Duffy, et~al.]{miikkulainen2019evolving}
Miikkulainen, R., Liang, J., Meyerson, E., Rawal, A., Fink, D., Francon, O.,
  Raju, B., Shahrzad, H., Navruzyan, A., Duffy, N., et~al.
\newblock Evolving deep neural networks.
\newblock In \emph{Artificial intelligence in the age of neural networks and
  brain computing}, pp.\  293--312. Elsevier, 2019.

\bibitem[Montanari \& Sen(2022)Montanari and Sen]{montanari2022short}
Montanari, A. and Sen, S.
\newblock A short tutorial on mean-field spin glass techniques for
  non-physicists.
\newblock \emph{arXiv preprint arXiv:2204.02909}, 2022.

\bibitem[Muralidhar et~al.(2018)Muralidhar, Islam, Marwah, Karpatne, and
  Ramakrishnan]{muralidhar2018incorporating}
Muralidhar, N., Islam, M.~R., Marwah, M., Karpatne, A., and Ramakrishnan, N.
\newblock Incorporating prior domain knowledge into deep neural networks.
\newblock In \emph{2018 IEEE international conference on big data (big data)},
  pp.\  36--45. IEEE, 2018.

\bibitem[Ng et~al.(2020)Ng, Cho, and Ghassemi]{ng2020ssmba}
Ng, N., Cho, K., and Ghassemi, M.
\newblock Ssmba: Self-supervised manifold based data augmentation for improving
  out-of-domain robustness.
\newblock \emph{arXiv preprint arXiv:2009.10195}, 2020.

\bibitem[Nguyen et~al.(2017)Nguyen, Zecchina, and Berg]{nguyen2017inverse}
Nguyen, H.~C., Zecchina, R., and Berg, J.
\newblock Inverse statistical problems: from the inverse ising problem to data
  science.
\newblock \emph{Advances in Physics}, 66\penalty0 (3):\penalty0 197--261, 2017.

\bibitem[Oreshkin et~al.(2019)Oreshkin, Carpov, Chapados, and
  Bengio]{oreshkin2019n}
Oreshkin, B.~N., Carpov, D., Chapados, N., and Bengio, Y.
\newblock N-beats: Neural basis expansion analysis for interpretable time
  series forecasting.
\newblock \emph{arXiv preprint arXiv:1905.10437}, 2019.

\bibitem[Ozaki(2012)]{ozaki2012time}
Ozaki, T.
\newblock \emph{Time series modeling of neuroscience data}.
\newblock CRC press, 2012.

\bibitem[Pan \& Yang(2009)Pan and Yang]{pan2009survey}
Pan, S.~J. and Yang, Q.
\newblock A survey on transfer learning.
\newblock \emph{IEEE Transactions on knowledge and data engineering},
  22\penalty0 (10):\penalty0 1345--1359, 2009.

\bibitem[Pan et~al.(2010)Pan, Ni, Sun, Yang, and Chen]{pan2010cross}
Pan, S.~J., Ni, X., Sun, J.-T., Yang, Q., and Chen, Z.
\newblock Cross-domain sentiment classification via spectral feature alignment.
\newblock In \emph{Proceedings of the 19th international conference on World
  wide web}, pp.\  751--760, 2010.

\bibitem[Pan et~al.(2023)Pan, Sharma, Hu, Liu, Li, Liu, Huang, and
  Geng]{pan2023Ising}
Pan, Z., Sharma, A., Hu, J. Y.-C., Liu, Z., Li, A., Liu, H., Huang, M., and
  Geng, T.~T.
\newblock Ising-traffic: Using ising machine learning to predict traffic
  congestion under uncertainty.
\newblock In \emph{Proceedings of the AAAI Conference on Artificial
  Intelligence}, 2023.

\bibitem[Raissi et~al.(2019)Raissi, Perdikaris, and
  Karniadakis]{raissi2019physics}
Raissi, M., Perdikaris, P., and Karniadakis, G.~E.
\newblock Physics-informed neural networks: A deep learning framework for
  solving forward and inverse problems involving nonlinear partial differential
  equations.
\newblock \emph{Journal of Computational Physics}, 378:\penalty0 686--707,
  2019.

\bibitem[Sezer et~al.(2020)Sezer, Gudelek, and Ozbayoglu]{sezer2020financial}
Sezer, O.~B., Gudelek, M.~U., and Ozbayoglu, A.~M.
\newblock Financial time series forecasting with deep learning: A systematic
  literature review: 2005--2019.
\newblock \emph{Applied Soft Computing}, 90:\penalty0 106181, 2020.

\bibitem[Sun et~al.(2015)Sun, Li, Liu, Chow, Sun, and Wang]{sun2015using}
Sun, Y., Li, J., Liu, J., Chow, C., Sun, B., and Wang, R.
\newblock Using causal discovery for feature selection in multivariate
  numerical time series.
\newblock \emph{Machine Learning}, 101\penalty0 (1):\penalty0 377--395, 2015.

\bibitem[Swendsen \& Wang(1987)Swendsen and Wang]{swendsen1987nonuniversal}
Swendsen, R. and Wang, J.
\newblock Nonuniversal critical dynamics in monte carlo simulations.
\newblock \emph{Physical Review Letters}, 58\penalty0 (2):\penalty0 86--88,
  1987.

\bibitem[Szegedy et~al.(2016)Szegedy, Vanhoucke, Ioffe, Shlens, and
  Wojna]{szegedy2016rethinking}
Szegedy, C., Vanhoucke, V., Ioffe, S., Shlens, J., and Wojna, Z.
\newblock Rethinking the inception architecture for computer vision.
\newblock In \emph{Proceedings of the International Conference on Computer
  Vision}, pp.\  2818--2826, 2016.

\bibitem[Tallec \& Ollivier(2018)Tallec and Ollivier]{tallec2018can}
Tallec, C. and Ollivier, Y.
\newblock Can recurrent neural networks warp time?
\newblock \emph{arXiv preprint arXiv:1804.11188}, 2018.

\bibitem[Tealab(2018)]{tealab2018time}
Tealab, A.
\newblock Time series forecasting using artificial neural networks
  methodologies: A systematic review.
\newblock \emph{Future Computing and Informatics Journal}, 3\penalty0
  (2):\penalty0 334--340, 2018.

\bibitem[Vaswani et~al.(2017)Vaswani, Shazeer, Parmar, Uszkoreit, Jones, Gomez,
  Kaiser, and Polosukhin]{vaswani2017attention}
Vaswani, A., Shazeer, N., Parmar, N., Uszkoreit, J., Jones, L., Gomez, A.~N.,
  Kaiser, {\L}., and Polosukhin, I.
\newblock Attention is all you need.
\newblock \emph{Advances in neural information processing systems}, 30, 2017.

\bibitem[V{\'a}zquez et~al.(2017)V{\'a}zquez, Del~Ferraro, and
  Ricci-Tersenghi]{vazquez2017simple}
V{\'a}zquez, E.~D., Del~Ferraro, G., and Ricci-Tersenghi, F.
\newblock A simple analytical description of the non-stationary dynamics in
  ising spin systems.
\newblock \emph{Journal of Statistical Mechanics: Theory and Experiment},
  2017\penalty0 (3):\penalty0 033303, 2017.

\bibitem[Weber \& Frey(2017)Weber and Frey]{weber2017master}
Weber, M.~F. and Frey, E.
\newblock Master equations and the theory of stochastic path integrals.
\newblock \emph{Reports on Progress in Physics}, 80\penalty0 (4):\penalty0
  046601, 2017.

\bibitem[Weiss et~al.(2016)Weiss, Khoshgoftaar, and Wang]{weiss2016survey}
Weiss, K., Khoshgoftaar, T.~M., and Wang, D.
\newblock A survey of transfer learning.
\newblock \emph{Journal of Big data}, 3\penalty0 (1):\penalty0 1--40, 2016.

\bibitem[Wen et~al.(2020)Wen, Sun, Yang, Song, Gao, Wang, and Xu]{wen2020time}
Wen, Q., Sun, L., Yang, F., Song, X., Gao, J., Wang, X., and Xu, H.
\newblock Time series data augmentation for deep learning: A survey.
\newblock \emph{arXiv preprint arXiv:2002.12478}, 2020.

\bibitem[Wissel(1979)]{wissel1979manifolds}
Wissel, C.
\newblock Manifolds of equivalent path integral solutions of the fokker-planck
  equation.
\newblock \emph{Zeitschrift f{\"u}r Physik B Condensed Matter}, 35\penalty0
  (2):\penalty0 185--191, 1979.

\bibitem[Yang et~al.(2015)Yang, Ravikumar, Allen, and Liu]{yang2015graphical}
Yang, E., Ravikumar, P., Allen, G.~I., and Liu, Z.
\newblock Graphical models via univariate exponential family distributions.
\newblock \emph{The Journal of Machine Learning Research}, 16\penalty0
  (1):\penalty0 3813--3847, 2015.

\bibitem[Yosinski et~al.(2014)Yosinski, Clune, Bengio, and
  Lipson]{yosinski2014transferable}
Yosinski, J., Clune, J., Bengio, Y., and Lipson, H.
\newblock How transferable are features in deep neural networks?
\newblock \emph{International Conference on Machine Learning}, pp.\
  3320--3328, 2014.

\bibitem[Zhou et~al.(2021)Zhou, Ma, and Liu]{zhou2021trade}
Zhou, Z., Ma, L., and Liu, H.
\newblock Trade the event: Corporate events detection for news-based
  event-driven trading.
\newblock \emph{arXiv preprint arXiv:2105.12825}, 2021.

\bibitem[Zhuang et~al.(2020)Zhuang, Qi, Duan, Xi, Zhu, Zhu, Xiong, and
  He]{zhuang2020comprehensive}
Zhuang, F., Qi, Z., Duan, K., Xi, D., Zhu, Y., Zhu, H., Xiong, H., and He, Q.
\newblock A comprehensive survey on transfer learning.
\newblock \emph{Proceedings of the IEEE}, 109\penalty0 (1):\penalty0 43--76,
  2020.

\end{thebibliography}
\bibliographystyle{icml2023}

\newpage

\onecolumn
\appendix

\part*{Appendix}

\begin{enumerate}[label=\Alph*]\setlength\itemsep{0em}
  \item \hyperref[appendix:related_works]{More Related Works}
  \item \hyperref[appendix:UEF_construction]{Theoretical Background: Constructing Multivariate Graphical Models from Univariate Exponential Family}
  \item \hyperref[appendix:proof]{Proof of Main Text}
  \item \hyperref[appendix:exp_details]{Experimental Details of Main Text}
  \item \hyperref[appendix:extra_exp]{Additional Experiments}
\end{enumerate}

\section{More Related Works}
\label{appendix:related_works}

Feature engineering is the process of constructing and selecting features for use in machine learning models. 
Compared to data augmentation \cite{ng2020ssmba,goodfellow2014generative,doersch2015unsupervised,szegedy2016rethinking}, which improves the generalization performance of the model by synthesizing data that is more similar to the test distribution \cite{bousmalis2016domain}, feature engineering is an important step in the modeling process focusing on extracting relevant and informative features from the raw data that can be used to train a model, as the quality of the features can have a significant impact on the performance of the model.  
Recent developments in feature engineering for machine learning include the use of automated techniques for feature selection and construction \cite{kaul2017autolearn,khurana2016cognito,guyon2003introduction}, the use of deep learning techniques for feature learning \cite{bengio2013representation, le2014distributed,hinton2012deep}, and the use of transfer learning techniques to adapt pre-trained feature representations to new tasks \cite{pan2009survey, yosinski2014transferable}. 

Among all these developments, automated feature engineering, the use of algorithms and computational tools to automatically construct and select features for use in machine learning models (including the use of genetic algorithms \cite{miikkulainen2019evolving} to evolve effective feature sets, the use of gradient-based optimization \cite{jiang2021efficient} to search for optimal feature combinations, and the use of deep learning techniques \cite{bengio2013representation, le2014distributed} for feature learning), is the primary interest of this paper --- as it allows for the handling of large and complex datasets, the discovery of complex feature interactions, and the rapid testing and comparison of different feature sets.

On the other hand, 
the field of automated feature engineering for time series modeling has also witnessed growing interest in recent times.
These methods usually function by extracting a significant number of predefined features from the data, followed by selecting a subset of the most pertinent ones, including univariate \cite{christ2018time}, multivariate time series \cite{de2022tsfuse} and many others mentioned in \cref{sec:intro}.

However, existing approaches may face limitations due to the issues highlighted in \cref{sec:intro}, such as the majority of features being either too domain-specific or not easily generated based on a fundamental principle. 
As a complement to these methods, our proposed programmable time series feature engineering framework provide a united perspective from a dynamical physics model, the spin-gas Glauber dynamics.
To the best of our knowledge, our framework represents the first theoretically grounded automated feature engineering method for time series, characterized by its comprehensiveness and physics-motivated fundamental principles.
Contrary to hand-crafted methods that typically rely on ad-hoc dictionaries and may struggle to transfer between different contexts, feature programming provides significant flexibility and adaptability across various tasks, domains, or applications through appropriate customization.

\section{Theoretical Background:\\Constructing Multivariate Graphical Models from Univariate Exponential Family}
\label{appendix:UEF_construction}

Here, we introduce multivariate graphical model construction using univariate exponential family distributions \cite{yang2015graphical}, which is used in  \cref{sec:theory} to pinpoint conditions enhancing the modeling capability of our proposed physics model in capturing complex time series.

Suppose $\bX= \(x_{1},\dots,x_{N}\)$ is a random vector of length $N$, with each random variable $x_{i}$ taking values in some set $\mathfrak{X}$.
Let $\calG=\{\calV, \calE\}$ be an undirected graph over the set of nodes $\calV\coloneqq [N]$ corresponding to $N$ variables $\{x_{i}\}_{i=1}^N$.
Let $\calC$ be the set of cliques (fully-connected subgraphs) of $\calG$, and $\{\Phi_c(\bX_c)\}_{c\in \calC}$ be the set of clique-wise sufficient statistics with $\bX_c$ denoting the variables within $\calC$.
Any distribution over $\bX$ within the graphical model family represented by $\calG$ has the exponential family form
\begin{align}
\label{eqn:graphical_model_family}
P(\bX) 
&\propto 
\exp{\sum_{c\in \calC} w_c\Phi_c(\bX_c)},
\end{align}
where $\{w_c\}_{c\in\calC}$ are the weights over $\{\Phi_c\}_{c\in \calC}$.
For $\calG$ being only up to pairwise cliques and $\mathfrak{X}$ being binary, \eqref{eqn:graphical_model_family} reduces to the Ising model family
$
P(\bX) \propto 
\exp{\sum_{i\in \calV} w_i \Phi(x_i) + \sum_{(j,k)\in \calE} w_{jk} \Phi(x_j,x_k)}
$.

Following \cite{yang2015graphical}, the graphical model \eqref{eqn:graphical_model_family} over multivariate $\bX$  can be constructed from a particular choice of univaraite exponential family for each random variable $x_{i}$ under conditions identified by Lemma~\ref{lemma:node_cond} \&~\ref{lemma:thm2} below.
\begin{definition}[Univariate Exponential Family]
    A univariate exponential family is a family of distributions for random variable $x$,
    \bea
    \label{eqn:uni_exp_family}
    P(x)= \exp{w \phi(x)+B(x_{i})-D(x)}, 
    \eea
    with $w$, $\phi(x_{i})$, $B(x_{i})$, $D(w)$ being the canonical exponential family parameter, \textit{univariate} sufficient statistics, base measure and log-partition function, respectively.
    The choice of $w$ and $\phi(x_{i})$ may vary depending on the particular distribution within the exponential family.
\end{definition}
Denoting $\calN(i)$ the set of neighbors of node $i$ according to $\calG$, we construct the node-conditionals based on above univariate exponential family distribution \eqref{eqn:uni_exp_family} by considering a potential function consisting of a linear combination of up to $k$-th order products of univariate sufficient statistics $\{\phi(x_j)\}_{j\in \calN(i)}$ and $\phi(x_i)$ :
\begin{align}
\label{eqn:node_conditional}
P\(x_i|\bX_{\calV{\setminus i}}\)
=\exp{\Psi_i\(\bX_{\calV\setminus i}\)\cdot \phi(x_i)+ B(x_i) - \Bar{D}\(\bX_{\calV{\setminus i}}\)
},
\end{align}
where $B(x_i)$ is specified by the univariate exponential family \eqref{eqn:uni_exp_family}, $\Bar{D}\(\bX_{\calV{\setminus i}}\)\coloneqq D\(\Psi_i\(\bX_{\calV\setminus i}\)\)$ denotes the conditional log-partition function, and the canonical parameter $\Psi_i\(\bX_{\calV\setminus i}\)$ of the univariate sufficient statistics function $\phi(x_i)$ is given by the following tensor-factorized form
\begin{align}
\label{eqn:canon_para_expand}
\Psi_i\(\bX_{\calV\setminus i}\)
=
w_i + \sum_{\alpha_2\in\calN(i)} w_{i\alpha_2} \phi(x_{\alpha_2}) 
+ \sum_{\alpha_2,\alpha_3\in\calN(i)}w_{i\alpha_2\alpha_3}\phi(x_{\alpha_2})\phi(x_{\alpha_3})
+\cdots+\sum_{\alpha_2,\dots,\alpha_k\in\calN(i)} w_{i\alpha_2\dots\alpha_k} \prod_{j=2}^k \phi(x_{\alpha_j}).
\end{align}
\begin{remark}
We remark that \eqref{eqn:canon_para_expand} linearly expands the canonical parameter $\Psi_i\(\bX_{\calV\setminus i}\)$ in terms of products of univaraite sufficient statistics of conditional variables $\{x_{\alpha_l}\}_{l\in \calV \setminus i}$ up to $(k-1)$-th order; and the node-conditional \eqref{eqn:node_conditional} is a univariate exponential family.
\end{remark}

By Hammersley-Clifford theorem \cite{clifford1990markov} and tenor factorization  \cite{yang2015graphical,besag1974spatial}, \eqref{eqn:node_conditional} can be shown to determine the joint distribution $P(\bX)$ uniquely:
    \begin{lemma}[Proposition~1 of \cite{yang2015graphical}]
    \label{lemma:node_cond}
    Given a graph $\calG$.
    Suppose $\bX=\(x_1,\dots,x_N\)$ is a random vector of size $N$ and its node-conditionals for each node is specified by a univariate exponential family \eqref{eqn:node_conditional} that factors according to $\calG$.
    Then its joint distribution sits inside the graphical model family, presented by $\calG$, of the form

    \begin{align}
    \label{eqn:joint_dist_from_node}
        P(\bX) =
        &\exp\Bigg\{
        \sum_{i\in\calV} w_i \phi(x_i)
        +\sum_{i\in\calV}\sum_{j\in\calN(i)} w_{ij} \phi(x_i)\phi(x_j)
        +\cdots
        +\sum_{i\in\calV}\cdots \sum_{\alpha_k\in\calN(i)}w_{i\alpha_2\dots\alpha_k} \phi(x_i)\prod_{l=2}^k \phi(\alpha_l)
        \nonumber\\
        &\quad\quad\quad+\sum_{i\in \calV} B(x_i)-A(\{w\})
        \Bigg\},
    \end{align}    
    where $A(\{w\})$ is the log-partition function of the joint distribution.
    \end{lemma}
Lemma~\ref{lemma:node_cond} states that,  if the node-conditional distributions are defined as a univariate exponential family given in \ref{eqn:node_conditional}, a unique graphical model (joint) distribution is specified by \ref{eqn:joint_dist_from_node}.
Interestingly, it can be also shown that, \eqref{eqn:node_conditional} and \eqref{eqn:joint_dist_from_node} are the most general form for pairwise graphical model by construction \cite{yang2015graphical}.

Reversely, we have the following lemma.
    \begin{lemma}[Theorem~2 of \cite{yang2015graphical}]
    \label{lemma:thm2}
        Suppose $\bX=\(x_1,\dots,x_N\)$ is a random vector of size $N$, whose node-conditionals are governed by the exponential family 
        \begin{align}
        \label{eqn:general_node_cond}
        P\(x_i|\bX_{\calV{\setminus i}}\)
        =\exp{\Theta\(\bX_{\calV \setminus i}\)\phi(x_i)+ B(x_i) - \Bar{D}\(\bX_{\calV{\setminus i}}\)
        },
        \end{align}
        with canonical parameter function $\Theta\(\bX_{\calV \setminus i}\)$,
        if the corresponding joint distribution $P(\bX)$ is a graphical model that factors according to some graph $\calG$ with clique-size at most $k$, then the conditionals  \eqref{eqn:general_node_cond} have the tensor-factorized form given by \eqref{eqn:node_conditional} (i.e. $\Theta\(\bX_{\calV \setminus i}\)=\Psi\(\bX_{\calV \setminus i}\)$)
        , and $P(\bX)$ takes the form of \eqref{eqn:joint_dist_from_node}.
    \end{lemma}
\cref{lemma:node_cond} and \cref{lemma:thm2} together specify the sufficient conditions for node-conditionals and joint distribution with respect to $\calG$ being given by \eqref{eqn:general_node_cond} and \eqref{eqn:joint_dist_from_node}, respectively: (i) node-conditionals are in a univariate exponential family, and (ii) the joint distribution is a graphical model factored according to $\calG$ with clique-size at most $k$.
This construction allows for several advantages, including the capability to model a wide variety of distributions, and closed-form multivariate densities that enable efficient inference and learning with statistical guarantees, see \cite{yang2015graphical} for details.

\section{Proof of Main Text}
\label{appendix:proof}

Our primary objective in the theory section is to demonstrate that the spin-gas Glauber dynamics introduced in \eqref{eqn:fine_grained_P} (or \eqref{eqn:thm_node_cond} for completeness) can be specified by a univariate exponential family and can be seen as the node-conditional of a multivariate graphical model $\calG$, which exhibits desirable properties according to \cite{yang2015graphical}.

We sketch the three conceptual steps here.
\begin{itemize}
    \item 
    \textbf{Step 1:} 
    We demonstrate that the node-conditional \eqref{eqn:thm_node_cond} is indeed specified by a univariate exponential family, and we identify the sufficient statistics and canonical parameters.
    \item 
    \textbf{Step 2:} 
    We show that the identified canonical parameters of \eqref{eqn:thm_node_cond} (and thus the spin-gas Glauber dynamics \eqref{eqn:fine_grained_P}) take on a tensor-factorized form if the effective local field $g$ is in the form of \eqref{eqn:g_tensor_factor}.
    \item 
    \textbf{Step 3:} 
    Finally, by applying \cref{lemma:node_cond} and \cref{lemma:thm2} with the tensor-factorized canonical parameters, we complete the proof, illustrating that our model inherits advantageous properties from \cite{yang2015graphical} under the identified conditions \eqref{eqn:g_tensor_factor}.
\end{itemize}

\subsection{Proof of Theorem~\ref{thm:main_thm}}
\label{proof:main_thm}

\begin{proof}
Recall $\calG=\(\calV_t,\calV_{t+\delta t},\calE\)$, and $\bX =\(\bm{\sigma}_t,\bm{\sigma}_{t+\delta t}\)$.
We first consider the node-conditionals given by the augmented Glauber dynamics, i.e. node-conditionals for nodes $q\in\calV_{t+\delta t}$.
Starting from expressing the augmented Glauber dynamics \eqref{eqn:fine_grained_P} into node-conditional form with 
canonical parameter $\Theta\(\bX_{\setminus (i,t+\delta t)}\)$, by assumption~\ref{assumption:evol_indept} (and consequently $\calE_{t+\delta t}=\emptyset$, $\calE_t\neq \emptyset$ and $\calE_{t,t+\delta t}\neq \emptyset$),
we have \eqref{eqn:fine_grained_P} taking node-conditional form 
\begin{align}
    P(\sigma_{i,t+\delta t}|\bm{\sigma}_t)
    =P(\sigma_{i,t+\delta t}|\bm{\sigma}_{t+\delta t\setminus i},\bm{\sigma}_t),
\end{align}
and therefore
\begin{align}
\frac{\exp{\sigma_{i,t+\delta t}\Gamma_{i,t}}}{2\cosh{\Gamma_{i,t}}}
&= \exp{\(\ln \half \cdot \Gamma_{i,t}\) \sigma_{i,t+\delta t} -\ln\[{\half \(e^{\Gamma_{i,t}}+e^{-\Gamma_{i,t}}+1\)}\]}
\nonumber\\
&=\exp{\Theta\(\bX_{\setminus (i,t+\delta t)}\)\Tilde{\phi}\(\sigma_{i,t+\delta t}\)+ B\(\sigma_{i,t+\delta t}\) - \Bar{D}\(\bm{\sigma}_t\)},\nonumber
\end{align}
where the last line takes the form of \eqref{eqn:general_node_cond}.
Here,  the $i$ index specifies a node in $\calV_{t+\delta t}$, and
the conditional potential function can be decomposed into two parts:
canonical parameter $\Theta\(\bX_{\setminus (i,t+\delta t)}\)=\ln \half \cdot \Gamma_{i,t}$ and sufficient statistics $\Tilde{\phi}\(\sigma_{t+\delta t}\)=\sigma_{i,t+\delta t} $.

From $\Theta(\bX_{\setminus (i,t+\delta t)})=\ln \half \cdot \Gamma_{i,t}$ and 
$\Tilde{\phi}(\sigma_{i,t+\delta t})=\sigma_{i,t+\delta t}$, we first observe that
\bea
\Theta(\bX_{\setminus (i,t+\delta t)})
=\Theta(\bX_{t+\delta t\setminus i}, \bm{\sigma}_t)
=\Theta_i(\bm{\sigma}_t),
\eea
where, we emphasize that, $\Theta_i(\bm{\sigma}_t)$ denotes the canonical parameter of the node-conditional of the $i$-th node in $\calV_{t+\delta t}$, and is a function of $\bm{\sigma}_t$ due to the definition of $\Gamma_{i,t}$.

For the sake of simplicity, we abuse notation by absorbing the $\ln \half$ factor into $\Gamma_{i,t}$, and write
\begin{align}
    \Theta_i(\bm{\sigma}_t)=\Gamma_{i,t}
    = 
    \sum_{j\in \calV_t} J_{ij}(t) \sigma_{j,t} + h_{i}(t) 
    +
    g_i\(\fdv{\bm{\sigma}_t}{t},\fdv{}{t}\(\fdv{\bm{\sigma}_t}{t}\) \),
\end{align}    
with $g_i(\cdot)$ satisfying \eqref{eqn:g_tensor_factor}.

Recall the fact that the univariate sufficient statistics functions $\{\Tilde{\phi}(\sigma_{\alpha_l})\}_{\alpha_l\in \calN(i)}$ are now functions of 0th-, 1st- and 2nd-order time derivatives, we expand the $\Psi_i\(\bX_{\setminus (i,t+\delta t)}\)$ using its definition
\begin{align}
    \Psi_i\(\bX_{t}\)=
    w_i + \sum_{\alpha_2\in\calN(i)} \Tilde{w}_{i\alpha_2} \Tilde{\phi}(\sigma_{\alpha_2}) 
    &+ \sum_{\alpha_2,\alpha_3\in\calN(i)}\Tilde{w}_{i\alpha_2\alpha_3}\Tilde{\phi}(\sigma_{\alpha_2})\Tilde{\phi}(\sigma_{\alpha_3})
    \nonumber\\
    &\quad\quad+\cdots+\sum_{\alpha_2,...,\alpha_k\in\calN(i)} \Tilde{w}_{i\alpha_2...\alpha_k} \prod_{j=2}^k \Tilde{\phi}(\sigma_{\alpha_j}).
\end{align}
We immediately read out $w_i=  h_{i}(t) $, 
and univariate sufficient statistics for node $\alpha_l\in \calN(i)\subseteq \calV_t$:
        \bea
        \Tilde{\phi}(\sigma_{\alpha_l})\coloneqq 
        \Tilde{\sigma}_{\alpha_l}^\sT= \(\sigma_{\alpha_l},\fdv{\sigma_{\alpha_l}}{t},\fdv{}{t}\(\fdv{\sigma_{\alpha_l}}{t}\)\)^\sT,
        \eea 
        with dimension of $\Tilde{w}$ being extended along each index (except $i$) by a factor of 3, 
        such that, for instance,
        \begin{align}
        \sum_{\alpha_2\in\calN(i)} \Tilde{w}_{i\alpha_2} \Tilde{\phi}(\sigma_{\alpha_2}) 
        &=\sum_{\alpha_2\in\calN(i)}
        \[w_{i\alpha_2} \sigma_{\alpha_2} +w'_{i\alpha_2}  \fdv{\sigma_{\alpha_2}}{t} +w''_{i\alpha_2} \fdv{}{t}\(\fdv{\sigma_{\alpha_2}}{t}\) \]
        \nonumber\\
        &=\sum_{j\in \calV_t} J_{ij}(t) \sigma_{j,t}
        + \sum_{j\in \calV_t} G^{(1)}_{ij} (t)\fdv{\sigma_{j,t}}{t}
        + \sum_{j\in \calV_t} G^{(2)}_{ij}(t) \fdv{}{t}\(\fdv{\sigma_{j,t}}{t}\).
        \end{align}
Higher-order product terms can be identified accordingly.
Therefore, we arrive the fact that the canonical parameter $\Theta_i(\bm{\sigma}_t)$ with $g_i(\cdot)$ given by \eqref{eqn:g_tensor_factor} is indeed in a tensor-factorized form $\Psi_i(\bm{\sigma}_t)$.
Moreover, the coefficients $\{\Tilde{w}\}$ can all be identified with couplings $\{G\}$.
We can therefore write down the extended sufficient statistics as a $(3N + 1N)$-dimensional vector  
\begin{align}
\Tilde{\phi}(\bX)
=\Big(\underbrace{\bigg\{\sigma_{i,t},\fdv{\sigma_{i,t}}{t},\fdv{}{t}\(\fdv{\sigma_{i,t}}{t}\)\bigg\}_{i=1}^N}_{3N},\underbrace{\{\sigma_{i,t+\delta t}\}_{i=1}^N}_N \Big)^{\sT}
\nonumber
\end{align}
and compactify the notation with index $\alpha_l\coloneqq (i,t)\in\calV_t$ and $\beta_l\coloneqq (i,t+\delta t)\in \calV_{t+\delta t}$,
\bea
\Tilde{\phi}(\bX)=\(\{\Tilde{\sigma}_{\alpha_l}\}_{l=1}^N, \{\sigma_{\beta_l}\}_{l=1}^N\)^\sT,
\eea
where, $
\Tilde{\phi}(\sigma_{\alpha_l})=\Tilde{\sigma}_{\alpha_l}\coloneqq \(\sigma_{\alpha_l},\fdv{\sigma_{\alpha_l}}{t},\fdv{}{t}\(\fdv{\sigma_{\alpha_l}}{t}\)\)$ for $\alpha_l \in \calV_t$, and $\Tilde{\phi}(\sigma_{\alpha_l})=\sigma_{\beta_l}$ $\beta_l \in \calV_{t+\delta t}$, are zero-padded vectors.

Consequently, we can determine the canonical parameter $\Theta_{i,t}$:
    \begin{align}
    \Psi_i\(\bX_{t}\)&=\Psi_i\(\bX_{t+\delta t\setminus i},\bm{\sigma}_t\)= \Gamma_{i,t}\(\bm{\sigma}_t\)
    \\
    &= 
    \sum_{j\in \calV_t} J_{ij}(t) \sigma_{j,t} + h_{i}(t) 
    +
    g_i\(\fdv{\bm{\sigma}_t}{t},\fdv{}{t}\(\fdv{\bm{\sigma}_t}{t}\) \)
    \nonumber\\
    &=
    w_i + \sum_{\alpha_2\in\calN(i)} w_{i\alpha_2} \Tilde{\phi}(\sigma_{\alpha_2}) + \sum_{\alpha_2,\alpha_3\in\calN(i)}w_{i\alpha_2\alpha_3}\Tilde{\phi}(\sigma_{\alpha_2})\Tilde{\phi}(\sigma_{\alpha_3})+\cdots
    \quad+\sum_{\alpha_2,...,\alpha_k\in\calN(i)} w_{i\alpha_2...\alpha_k} \prod_{j=2}^k \Tilde{\phi}(\sigma_{\alpha_j}),\nonumber
    \end{align}
    where  $\alpha_l\in \calN(i)\subseteq \calV_t$ and univariate functions $\{\Tilde{\phi}(\sigma_{\alpha_l})\}_{\alpha_l\in \calN(i)}$ are functions of 0th-, 1st- and 2nd-order time derivatives.

    Finally, we observe that, since the node-conditionals of nodes $q\notin \calV_{t+\delta t}$ follow the standard Glauber dynamics and $G_{t,t+\delta t}$ consists cliques of at most size $k$, the node-conditionals given by \eqref{eqn:thm_node_cond} and \eqref{eqn:completeness_cond} satisfy the conditions of Lemma~\ref{lemma:thm2} and hence \eqref{eqn:thm_node_cond} indulges the tensor-factorized form \eqref{eqn:g_tensor_factor}. 
    Therefore, we complete the proof by applying Lemma~\ref{lemma:node_cond} and determining joint distribution $P(\bX)$ of $\calG_{t,t+\delta t}$ uniquely.
\end{proof}

\clearpage
\section{Experimental Details  of Main Text}
\label{appendix:exp_details}

\subsection{Datasets}
\label{appendix:data}
We exploit four supervised regression datasets  to test the proposed framework.
\begin{itemize}
    \item \textbf{Synthetic Dataset:}
    To validate our method, we employ a synthetic dataset designed to align with the method's underlying assumptions. We derive this dataset from the Taxi Dataset, replacing the original input features with a mixture of zero-order, first-order, and second-order features, essentially omitting the basic features. For the output values, we opt for a randomly selected subset of zero-order features, such as volatility or exponential smoothing, derived from the foundational basic features, forecasting one step ahead for all input series.
    \item  \textbf{Taxi Dataset:}
    We use the TLC Trip Record Dataset, which has the number of taxi rides for 1000 locations in the form of 30 minute time intervals. We use the current 30-minute interval to forecast the next 30-minute interval.

    \item  \textbf{Electricity Dataset:}
    We use the UCI Electricity Load Diagrams Dataset, which has the electricity consumption of 370 customers in the form of hourly time intervals. 
    We use the current hour to forecast the next hour.      

    \item  \textbf{Traffic Dataset:}
    We use the UCI PEM-SF Traffic Dataset holds the occupancy rate of 440 San Francisco Bay Area highways in the form of hourly time intervals. We use the current hour to forecast the next hour.

\end{itemize}

\paragraph{Extended Features.}
    Let $N$ be the number variate in the multivariate time series, $K$ be the number of extended features and $T$ be the sequence length.
    We use the default setting of the feature program to generate $K=45$ features from each dataset (which has one feature for each variate).
    Namely, the data formats of the basic and extended features are $(N,1,T)$ and $(N,K,T)$, respectively.

    To generate the extended features, we utilize all of the fundamental operators on the output of the previous order at each order's feature calculation. 
    For the 0th order features, we applied smoothing using the window operator with lookback size of  [7, 25]. For the 1st order features, we applied the window, difference, and shift operators with lookback size of [7, 25]. For the 2nd order features, we applied the window, difference, and shift operators with lookback size of [7, 25]. Our final extended features set is composed of the basic features and the all of the generated features at each order.

\paragraph{Noisiness.}
To see the noisiness of each dataset, we compute the Temporal Signal-to-Noise Ratio (TSNR) for all datasets.
\begin{table}[h]
\vspace{-0.2truein}
    \centering
    \caption{Assessment of Dataset Noisiness using Temporal Signal-to-Noise Ratio (TSNR) for each Dataset. 
    The taxi dataset emerges as the noisiest, and our experimental findings highlight the increased significance of extended features in handling such noisy datasets.
    }
    \vspace*{0.05truein}
    \begin{tabular}{cccccc}
    \toprule
    \toprule
        & \multicolumn{1}{c}{Synthetic} 
         & \multicolumn{1}{c}{Taxi} 
         & \multicolumn{1}{c}{Electric} 
         & \multicolumn{1}{c}{Traffic} 
         \\ 
        \midrule
        \makecell{TSNR} 
        & 1.60
        & 1.27 
        & 2.98
        & 1.32
        \\ \bottomrule\bottomrule
    \end{tabular}
    \label{table:noise}
\end{table}

\subsection{Hyperparameter Search}
Hyperparameter optimization is conducted via
random search for 100 iterations.
\begin{itemize}
    \item $\mathtt{learning\_rate}$: 0.01, 0.001, 0.0001, 0.00001
    \item $\mathtt{batch\_size}$: 64, 128, 256, 512, $\mathtt{feature\_dim}$
    \item $\mathtt{hidden\_size}$: 64, 128, 512, 1024, 2048
    \item $\mathtt{num\_epochs}$: we use early stopping.
\end{itemize}

\subsection{Implementation Details }
\label{appendix:implementation}
For each experimental setting, we set the batch size to equal the number of variables for the corresponding dataset.
\begin{itemize}
    \item 
    \textbf{Architectural Details:}
    For the MLP model, the architecture is composed of 3 layers, with a hidden size of 512, and ReLU activation function. 
    For the CNN model, the architecture is composed of 1 fully connected layer with hidden size of 512 and ReLU activation function, and 2 convolutional layers with kernel size of 3. 
    For the LSTM model, the architecture is composed of 2 hidden layers and a hidden size of 512.
    \item 
    \textbf{Training Details:}
            We use an Adam optimizer with learning rate $\mathtt{lr}=10^{-5}$ for training. 
            The coefficients of Adam optimizer, betas, are set to $(0.9,0.999)$. 
        \item 
        \textbf{Platforms:}
        The GPUs and CPUs used to conduct experiments are NVIDIA GEFORCE RTX 2080 Ti and INTEL XEON SILVER 4214 @ 2.20GHz.
\end{itemize}

\section{Additional Experiments}
\label{appendix:extra_exp}
To complement the experiments presented in \cref{sec:exp}, we carry out further investigations that include more advanced models, practical scenarios, illustrative examples of customizing the feature program, and ablation studies to demonstrate the impact of operators on the generated features.

\subsection{Evaluating Feature Programming Across Diverse Deep Learning Models}
\label{appendix:many_models}
We conduct further experiments using a variety of widely-used time series forecasting models, with an emphasis on standard one-step-ahead predictions. 
In addition, we also incorporate multi-horizon prediction task experiments to address practical scenarios.

\paragraph{Models.} 
\begin{itemize}
    \item \textbf{XGBoost} \cite{chen2015xgboost}
    \item \textbf{LightGBM} \cite{ke2017lightgbm}

    \item \textbf{Transformer} \cite{vaswani2017attention}
    \item \textbf{Temporal Fusion Transformer (TFT)} 
    \cite{lim2021temporal}
    \item \textbf{Temporal Convolution Network (TCN)} \cite{chen2020probabilistic}
    \item \textbf{N-BEATS} \cite{oreshkin2019n}
\end{itemize}
We utilize the \code{DART}\footnote{\href{https://unit8co.github.io/darts/}{\texttt{https://unit8co.github.io/darts/}}} package to implement  XGBoost, LightGBM, Transformer, TFT, TCN, and N-BEATS.

\paragraph{Data.} 
We employ the easiest dataset and the most challenging dataset identified in \cref{sec:exp} and specified in \cref{appendix:data} --- the synthetic and taxi datasets.

\paragraph{Experiment Settings.}
For each of the datasets, we do an 80/20 train/test split to get our training data (in-sample) and testing data (out-of-sample).
For all experiments, we conduct 5 runs and report the average performance.
For the models, we stick to the common default configurations used in literature so as for a fair comparison.

\paragraph{Problem Setting: One-Step-Ahead Prediction with Extended Features.}
We conduct one-step-ahead predictions using above models and assess their performance by comparing results obtained with and without the inclusion of extended features.
We examine the standard one-step-ahead time series regression problem with a lookback size of $T=20$, which corresponds to a sequence length of $20$.
As illustrated in Table~\ref{table:results_many_model_1ahead}, our extended features yield consistent performance enhancements in prediction tasks when used as model inputs for a range of machine learning (XGBoost and LightGBM) and deep neural network time series prediction models (Transformer, TCN, TFT and N-BEATS).

\begin{table*}[h]
    \centering
    \caption{Performance Comparison of Common Time Series Models with and without Extended Features on Synthetic and Taxi Datasets. The table demonstrates the performance enhancements consistently achieved in one-step-ahead prediction tasks using a lookback size of $T=20$ when extended features are utilized as inputs for various machine learning (XGBoost and LightGBM) and deep neural network time series prediction models (Transformer, TCN, TFT and N-BEATS).}
    \vspace*{0.05truein}
    \resizebox{\textwidth}{!}{
        \begin{tabular}{clcccccccccccc}
        \toprule
        \toprule
            \multirow{2}{*}{Metric}
            &\multirow{2}{*}{Dataset}  
            & \multicolumn{2}{c}{XGBoost} 
            & \multicolumn{2}{c}{LightGBM} 
            & \multicolumn{2}{c}{Transformer} 
            & \multicolumn{2}{c}{TCN} 
            & \multicolumn{2}{c}{TFT} 
            & \multicolumn{2}{c}{N-BEATS} \\ 
            \cmidrule{3-14}
            & & Basic & Extended
            & Basic & Extended
            & Basic & Extended
            & Basic & Extended
            & Basic & Extended
            & Basic & Extended \\
            \midrule
            \multirow{2}{*}{$R^2$ Score \%} 
            & {\small Synthetic} 
            & 99.12 & \textbf{99.34} 
            & 98.92 & \textbf{99.34} 
            & 96.46 & \textbf{96.86} 
            & 98.87 & \textbf{99.37} 
            & 96.21 & \textbf{97.60} 
            & 99.26 & \textbf{99.41} \\
            & {\small Taxi} 
            & 77.34 & \textbf{78.15}
            & 77.05 & \textbf{81.01}
            & 72.64 & \textbf{73.70}
            & 75.36 & \textbf{77.70}
            & 54.58 & \textbf{61.31} 
            & 76.82 & \textbf{79.94} \\
            \midrule
            \multirow{2}{*}{\makecell{Pearson \\ Correlation} \%}
            & {\small Synthetic} 
            & 99.56 & \textbf{99.67}   
            & 99.46 & \textbf{99.67} 
            & \textbf{99.11} & 99.08 
            & 99.44 & \textbf{99.69} 
            & 98.41 & \textbf{99.19}  
            & 97.63 & \textbf{99.73} \\
            & {\small Taxi} 
            & 87.96 & \textbf{88.42} 
            & 87.80 & \textbf{88.97}
            & 85.54 & \textbf{86.83}
            & 86.87 & \textbf{88.26}
            & 76.65 & \textbf{79.65} 
            & 87.84 & \textbf{89.61}  \\
            \bottomrule
            \bottomrule
        \end{tabular}
        \label{table:results_many_model_1ahead}
    }
\end{table*}

\paragraph{Multi-Horizon Prediction with Extended Features.}
We further investigate multi-horizon prediction tasks with horizon sizes of 1, 2, 3, 5, 10 and 20, utilizing a lookback size of $T=20$. 
The models applied for multi-horizon prediction include Transformer, TFT, TCN\footnote{The current version of \code{DART} package requires that the output length is strictly smaller than the input length. 
As a result, we are unable to report the 20-step horizon results for TCN. Instead, we have included the results for a 19-step horizon and labeled them with $^*$ in \cref{table:multi-horizon}.}, N-BEATS. 
These experiments are conducted using the synthetic dataset, and reported in Table~\ref{table:multi-horizon}.
The results showcase the effectiveness of the generated features in various multi-horizon prediction settings.
Particularly, when forecasting the next 20 values using a sequence of length 20 (arguably the toughest task when only relying on basic features), the features generated by our method have demonstrated significant improvements in prediction performance across all models, suggesting that the extended features possess a greater amount of autoregressive information compared to basic features.

\begin{table*}[h]
    \centering
    \caption{Evaluation of Feature Quality for Multi-Horizon Prediction using Synthetic Dataset. This table presents the performance of the Transformer, TFT, TCN and N-BEATS models on multi-horizon prediction tasks with horizon sizes of 1, 2, 3, 5, 10 and 20, benchmarked on the synthetic dataset. The results showcase the effectiveness of the generated features in various multi-horizon prediction settings. Particularly, when predicting a full-length sequence (using a sequence of length 20 to predict the next 20 values), our generated features have demonstrated considerable prediction improvements across all models, suggesting that the extended features possess a greater amount of autoregressive information compared to basic features.
    }
    \vspace*{0.05truein}
    \resizebox{ \textwidth}{!}{
        \begin{tabular}{cccccccccccccc}
        \toprule
        \toprule
        \multirow{2}{*}{Model}&\multirow{2}{*}{Metric}& \multicolumn{2}{c}{1-Step Horizon}
        & \multicolumn{2}{c}{2-Step Horizon}
        & \multicolumn{2}{c}{3-Step Horizon}
        & \multicolumn{2}{c}{5-Step Horizon}
        & \multicolumn{2}{c}{10-Step Horizon}
        & \multicolumn{2}{c}{20-Step Horizon} \\
        \cmidrule{3-14}
        & & Basic & Extended & Basic & Extended & Basic & Extended & Basic & Extended & Basic & Extended & Basic & Extended \\ 
        \midrule
        \multirow{2}{*}{Transformer}
        & {\small $R^2$ Score \%} 
        & 96.46 & \textbf{96.86}
        & 94.47 & \textbf{96.58}
        & 94.07 & \textbf{96.92}
        & 91.03 & \textbf{92.78}
        & 73.11 & \textbf{81.17}
        & 54.46 & \textbf{73.33} \\ 
        & {\small Pearson Correlation \%}
        & \textbf{99.11} & 99.08 
        & 98.02 & \textbf{98.63}
        & 97.11 & \textbf{98.48}
        & 95.42 & \textbf{96.40}
        & 85.76 & \textbf{90.41}
        & 74.31 & \textbf{86.20} \\
        \midrule
        \multirow{2}{*}{TFT}
        & {\small $R^2$ Score \%} 
        & 96.21 & \textbf{97.60}
        & 95.81 & \textbf{96.86}
        & 93.99 & \textbf{95.11}
        & 84.76 & \textbf{92.29}
        & 60.87 & \textbf{78.60}
        & 35.35 & \textbf{64.91} \\
        & {\small Pearson Correlation \%}
        & 98.41 & \textbf{99.19}
        & 97.92 & \textbf{98.57}
        & 97.30 & \textbf{98.02}
        & 93.39 & \textbf{96.11}
        & 81.14 & \textbf{88.75}
        & 66.13 & \textbf{80.85} \\
        \midrule
        \multirow{2}{*}{TCN}
        & {\small $R^2$ Score \%} 
        & 98.87 & \textbf{99.37}
        & 96.03 & \textbf{97.98}
        & 92.41 & \textbf{95.75}
        & 79.91 & \textbf{88.32}
        & 41.91 & \textbf{66.18}
        & 17.18$^*$ & \textbf{52.50}$^*$ \\
        & {\small Pearson Correlation \%}
        & 99.44 & \textbf{99.69}
        & 98.02 & \textbf{99.01}
        & 96.20 & \textbf{97.86}
        & 89.51 & \textbf{94.11}
        & 65.51 & \textbf{81.41}
        & 46.08$^*$ & \textbf{73.19}$^*$ \\
        \midrule
        \multirow{2}{*}{N-BEATS}
        & {\small $R^2$ Score \%} 
        & 99.26 & \textbf{99.41}
        & 98.39 & \textbf{98.87}
        & 97.26 & \textbf{98.00}
        & 93.17 & \textbf{95.69}
        & 78.09 & \textbf{86.97}
        & 57.06 & \textbf{74.65} \\
        & {\small Pearson Correlation \%}
        & 97.63 & \textbf{99.73}
        & 99.20 & \textbf{99.44}
        & 98.62 & \textbf{99.07}
        & 96.62 & \textbf{97.84}
        & 88.90 & \textbf{93.28}
        & 75.80 & \textbf{86.44} \\
        \bottomrule
        \bottomrule
        \end{tabular}
    \label{table:multi-horizon}
}
\end{table*}

\paragraph{Simple Models Using Extended Features Achieve Comparable Performance to SOTA Models.}
As demonstrated in Table~\ref{table:results_many_model_1ahead} and Table~\ref{table:multi-horizon},  utilizing extended features as inputs for simple models (e.g. Transformer) allows them to reach performance levels comparable to state-of-the-art models (TCN, TFT, N-BEATS) that rely solely on raw features.

\subsection{Customizing Feature Programs: Three Examples of Customized Feature Programs}
\label{appendix:customization}

Our framework forms the crux of customizable feature engineering for multivariate time series. It's highly adaptable and doesn't depend on the model's structure. The production of extended features depends solely on the custom design of the pre-programmed operation module and the feature template. This proves especially advantageous when deep learning models struggle to extract specific features or when users wish to ensure that a certain aspect of the data is considered by the model.
In this section, we present three illustrative examples to showcase the customization procedure that may be required in real-world scenarios. 
These examples show how to tailor the feature program to resemble known hand-crafted features.
This corresponds to common practical scenarios where we have some insights into the types of features that might be particularly informative for the specific task, domain, or application at hand. 
For demonstration purposes, we adopt a simplified setting that employs straightforward and commonly known hand-crafted features (Momentum $\mathtt{MoM}$, Bias $\mathtt{Bias}$ and Absolute Energy $\mathtt{AbsEnergy}$) as the targets features for resembling.

\paragraph{Momentum $\mathtt{MoM}[t,\Delta \tau]$.}

As mentioned in \cite{gu2020empirical}, one of the simplest common hand-crafted feature for time series is the momentum feature, which refers to the rate of change in a time series over a specified period. 
It is a measure of the series recent performance, which is often used to identify trending time series that are expected to continue their movements in the same direction. 

Let $\Delta \tau$ be the lookback size of the momentum feature.
The momentum feature is defined as:
\bea
\mathtt{MoM}[t,\Delta \tau] = \frac{x_{t} - x_{t-\Delta \tau}}{x_{t-\Delta \tau}}.
\eea
It calculates the percentage change in the time series values over the past $\Delta \tau$ time steps. 
A positive momentum value indicates that the time series value has increased during that period, while a negative value means the value has decreased.

To resemble $\mathtt{MoM}[t,\Delta \tau]$ with the feature programming framework, we set the feature template and operation module as follows: 
\begin{enumerate}
    \item Set feature template:
    \begin{enumerate}
        \item Set both \code{0th basic feature}, and \code{2nd basic feature} to be empty.
        \item Incorporate \code{ratio feature}, defined as
        $
        \mathtt{ratio}[a,b]\coloneqq a/b 
        $,
        into \code{1st basic feature}.
    \end{enumerate}

    \item Pre-program the operation module:
    \begin{itemize}
        \item \textbf{0th order:}
        \begin{enumerate}
        \item  Perform \code{Shift[$x_t,\Delta\tau$]} and obtain $\{x_{t-\Delta \tau}\}$.
        \item  Perform \code{Difference[$x_t,x_{t-\Delta\tau}$]} and obtain 1st-order series $\{x_{t} - x_{t-\Delta \tau}\}$. 
        \end{enumerate}

    \item \textbf{1th order}:
    \begin{enumerate}
        \item Pass $a=\{x_{t-\Delta \tau}\}$ and $b=\{x_{t}-x_{t-\Delta \tau}\}$ as \code{1st basic feature} (where $\mathtt{ratio}[a,b]$ is in.)
        \item The output from \code{1st basic feature} gives the resembled $\mathtt{MoM}[t,\Delta \tau]$.
    \end{enumerate}
    \end{itemize}
\end{enumerate}
It is important to note that the computational trajectory presented here specifically focuses on the output feature that resembles $\mathtt{MoM}[t,\Delta \tau]$. 
The final set of extended features may include other distinct extended features.

\paragraph{BIAS $\mathtt{Bias}\[t, \Delta \tau \]$.}
The second example we considered is the bias feature, which indicated the current trend of the time series and the potential reversals.
Let $\mathtt{SMA}[t,\Delta \tau]\coloneqq \sum_{i=t-\Delta\tau+1}^{t}x_i/\Delta\tau$ be the sample moving average of lookback size $\Delta \tau$. 
The bias feature is defined as
    \bea 
    \mathtt{Bias}\[t, \Delta \tau \]\coloneqq
    \frac{x_t-\mathtt{SMA}[t,\Delta \tau]}{\mathtt{SMA}[t,\Delta \tau]}.
    \eea 
To resemble $\mathtt{Bias}\[t, \Delta \tau, \]$, we set the feature template and operation module as follows: 
\begin{enumerate}
    \item Set feature template:
    \begin{enumerate}
        \item Incorporate \code{SMA feature}, defined as
        $\mathtt{SMA}[t,\Delta \tau]
        $,
        into \code{0th basic feature}.
        \item Incorporate \code{ratio feature} 
        into \code{1st basic feature}.
        \item Set \code{2nd basic feature} to be empty.
    \end{enumerate}

    \item Pre-program the operation module:
    \begin{itemize}
        \item \textbf{0th order:}
        \begin{enumerate}
        \item  Perform \code{Difference[$x_t,\mathtt{SMA}[t,\Delta \tau]$]} and obtain 1st-order series $\{x_{t} - \mathtt{SMA}[t,\Delta \tau]\}$. 
        \end{enumerate}

    \item \textbf{1th order}:
    \begin{enumerate}
        \item Pass $a=\{x_t,\mathtt{SMA}[t,\Delta\tau]\}$ and $b=\{\mathtt{SMA}[t,\Delta\tau]\}$ as \code{1st basic feature} (where $\mathtt{ratio}[a,b]$ is in.)
        \item The output from \code{1st basic feature} gives the resembled $\mathtt{Bais}[t,\Delta \tau]$.
    \end{enumerate}
    \end{itemize}
\end{enumerate}

\paragraph{Absolute Energy $\mathtt{AbsEnergy}[t,\Delta\tau]$.}
Following \cite{christ2018time}, the third example we considered is the absolute energy feature which is often used in signal processing, particularly for identifying the energy content of a signal, and it can also be useful in time series analysis for understanding the overall strength or magnitude of the data. 
The absolute energy feature  is defined as
\bea
\mathtt{AbsEnergy}[t,\Delta\tau]\coloneqq \sum_{i=t-\Delta\tau+1}^t x_i^2.
\eea
To resemble $\mathtt{AbsEnergy}[t,\Delta \tau]$, we set the feature template and operation module as follows: 
\begin{enumerate}
    \item Set feature template:
    \begin{enumerate}
        \item Incorporate \code{square feature}, $\mathtt{square}[a]\coloneqq a^2$, 
        into \code{0th basic feature}.
        \item Incorporate \code{sum feature}, $\mathtt{sum}[a_t,\Delta \tau]\coloneqq \sum_{i=t-\Delta\tau+1}^{t}a_i$, 
        into \code{1st basic feature}.
        \item Set \code{2nd basic feature} to be empty.
    \end{enumerate}

    \item Pre-program the operation module:
    \begin{itemize}
        \item \textbf{0th order:}
        \begin{enumerate}
        \item  Do nothing.
        \end{enumerate}

    \item \textbf{1th order}:
    \begin{enumerate}
        \item Pass $a=\{x^2_{t}\}$ (from \code{0th basic feature}) as \code{1st basic feature} (where $\mathtt{sum}[a_t,\Delta\tau]$ is in.)
        \item The output from \code{1st basic feature} gives the resembled $\mathtt{AbsEnergy}[t,\Delta \tau]$.
    \end{enumerate}
    \end{itemize}
\end{enumerate}

\paragraph{Evaluating the Resemblance of Hand-Crafted Features.}
From the results above, by construction, we can achieve an \textit{exact} resemblance of the given hand-crafted features by appropriately customizing the feature program. 
Following the same procedure, more complex features can also be programmed and resembled.
Generally, to evaluate the quality of the resembled features, we follow the feature generation flow (path in the computational graph) and identify the specific feature output that resembles the target hand-crafted features. 
In our case, we have $\mathtt{MoM}[t,\Delta \tau]$, $\mathtt{Bais}[t,\Delta \tau]$, and $\mathtt{AbsEnergy}[t,\Delta \tau]$ all perfectly resembled, with both $R^2$ and Pearson correlation equal to $1$ when comparing the resembled features to the hand-crafted features. 
For the comparison baselines, we use $\mathtt{MoM}$ from \cite{gu2020empirical}, $\mathtt{Bias}$ from the \code{PandasTA}\footnote{\href{https://github.com/twopirllc/pandas-ta}{\texttt{https://github.com/twopirllc/pandas-ta}}} package, and $\mathtt{AbsEnergy}$ from \cite{christ2018time}.

We would like to stress that achieving such exact resemblance is not generally applicable in practice (as it requires prior knowledge of the features to use), but the customization process for the feature program is indeed general. It is always possible to program desired (hand-crafted) features as long as they can be represented as programmable functions.

\end{document}